\documentclass[runningheads]{llncs}
\usepackage[T1]{fontenc}
\usepackage{tikz}
\usetikzlibrary{calc}
\usepackage{booktabs}
\usepackage{multirow}
\usepackage{makecell}
\usepackage{ragged2e}
\usepackage{comment}
\usepackage{subcaption}
\usepackage[numbers]{natbib}
\usepackage{graphicx}
\usepackage{amsmath}
\begin{document}
\title{Investigating Zero-Shot Diagnostic Pathology in Vision-Language Models with Efficient Prompt Design}

\author{Vasudev Sharma\inst{1}\textsuperscript{\dag} \and
Ahmed Alagha \inst{1}\textsuperscript{\dag} \and
Abdelhakim Khellaf \inst{2} \and \\
Vincent Quoc-Huy Trinh \inst{2} \and
Mahdi S. Hosseini\inst{1,3}}

\institute{Department of Computer Science and Software Engineering (CSSE) Concordia University, Montreal, Canada \and University of Montreal Hospital Center, Montreal, Canada \and Mila–Quebec AI Institute, Montreal, Canada}

%\author{First Author\inst{1}\orcidID{0000-1111-2222-3333} \and
%Second Author\inst{2,3}\orcidID{1111-2222-3333-4444} \and
%Third Author\inst{3}\orcidID{2222--3333-4444-5555}}
%\institute{Princeton University, Princeton NJ 08544, USA \and
%Springer Heidelberg, Tiergartenstr. 17, 69121 Heidelberg, Germany
\maketitle              % typeset the header of the contribution

\renewcommand{\thefootnote}{\dag}
\footnotetext{Equal contribution.}
\renewcommand{\thefootnote}{\arabic{footnote}} % reset footnote style if needed

\begin{abstract}
Vision-language models (VLMs) have gained significant attention in computational pathology due to their multimodal learning capabilities that enhance big-data analytics of giga-pixel whole slide image (WSI). However, their sensitivity to large-scale clinical data, task formulations, and prompt design remains an open question, particularly in terms of diagnostic accuracy. In this paper, we present a systematic investigation and analysis of three state of the art VLMs for histopathology, namely Quilt-Net, Quilt-LLAVA, and CONCH, on an in-house digestive pathology dataset comprising 3,507 WSIs, each in giga-pixel form, across distinct tissue types. Through a structured ablative study on cancer invasiveness and dysplasia status, we develop a comprehensive prompt engineering framework that systematically varies domain specificity, anatomical precision, instructional framing, and output constraints. Our findings demonstrate that prompt engineering significantly impacts model performance, with the CONCH model achieving the highest accuracy when provided with precise anatomical references. Additionally, we identify the critical importance of anatomical context in histopathological image analysis, as performance consistently degraded when reducing anatomical precision. We also show that model complexity alone does not guarantee superior performance, as effective domain alignment and domain-specific training are critical. These results establish foundational guidelines for prompt engineering in computational pathology and highlight the potential of VLMs to enhance diagnostic accuracy when properly instructed with domain-appropriate prompts.

\keywords{Vision-Language Models \and Computational Pathology \and Prompt Engineering \and Histopathology \and CONCH \and Quilt-Net \and Quilt-LLAVA}
\end{abstract}

\section{Introduction}
Computational pathology has witnessed a significant growth with the recent advancements of artificial intelligence (AI). Many deep learning (DL) models, particularly convolutional neural network (CNNs) and transformer architectures, have been developed in the past decade with remarkable success in computational pathology and whole slide image (WSI) analysis for tasks like disease diagnosis and prognosis \cite{Cui2021-jd, hosseini2024computationalpathologysurveyreview}. WSIs are giga-pixel images scanned from tissue microscopy scanners in digital pathology  that require to be processed for obtaining high-relevance diagnostic information. With the expansion of high-throughput digital pathology and the accumulation of vast repositories of WSIs, vision-language models (VLMs) came as powerful tools that leverage big data to enhance diagnostic accuracy and interpretability \cite{lu2023visuallanguagefoundationmodelcomputational}. Their multi-modal learning capabilities that integrate textual and visual information, where image-based features and domain-specific language are combined, have shown great success in addressing several pathological tasks, including tumor classification, tissue segmentation, and disease prognosis \cite{ikezogwo2025quilt1mmillionimagetextpairs, lu2023visuallanguagefoundationmodelcomputational}. This integration addresses the fundamental challenges in pathology where expert knowledge is encoded in natural language reports while diagnostic evidence is captured in visual data. Recent advances in self-supervised learning \cite{caron2021emergingpropertiesselfsupervisedvision, oquab2024dinov2learningrobustvisual} and transformer architectures \cite{vaswani2023attentionneed, dosovitskiy2021imageworth16x16words} have catalyzed the development of VLMs that can effectively bridge these modalities. Nonetheless, the sensitivity of these models to task complexity, data variations, and prompt design remains an open question.

Within the realm of VLMs in computational pathology, prompt variation has emerged as a critical factor, where even subtle modifications in input phrasing can yield different prediction outcomes \cite{hartsock2024prompt}. Notably, several recent works have proposed dynamic prompt optimization techniques, such as interpretable prompt optimization and attribute-guided prompt tuning to mitigate this sensitivity and enhance robustness on novel classes \cite{du2024IPO, gu2023robust}. To further address these issues, researchers are developing adaptive prompt tuning methods, including reinforcement learning from human feedback \cite{ouyang2022training}, that dynamically adjust prompts to stabilize model responses and enhance prediction reliability.

This study systematically investigates and evaluates the performance of three state-of-the-art VLM-based methods for pathology, namely Quilt-Net \cite{ikezogwo2025quilt1mmillionimagetextpairs}, Quilt-LLAVA \cite{seyfioglu2025quiltllavavisualinstructiontuning}, and CONCH \cite{lu2023visuallanguagefoundationmodelcomputational}. Quilt-Net is an instance of the contrastive language-image pre-training (CLIP) model \cite{radford2021learningtransferablevisualmodels}, which is fine-tuned on a large-scale dataset of 1M histopathology image-text pairs. Quilt-LLAVA extends Quilt-Net by adopting the large language and vision assistant (LLAVA) large multimodal model (LMM) framework \cite{liu2023visualinstructiontuning}, which enhances the contextual reasoning and prompt adaptability of the model. Contextual reasoning in Computational Histopathology (CONCH) is a vision-language foundation model trained on 1.17M histopathology image-caption pairs through task-agnostic pre-training based on the Contrastive Captioners (CoCa) method \cite{yu2022cocacontrastivecaptionersimagetext}. Using an in-house dataset of clinically validated 3507 digestive giga-pixel WSIs, we assess these models in terms of diagnostic accuracy while analyzing their sensitivity to prompt engineering that varies domain specificity, anatomical precision, instructional framing, and output constraints. Specifically, this study has the following contributions:

\begin{enumerate}
    \item We conduct a comprehensive ablative study to understand the effects of different prompt design strategies on different VLMs capabilities.
    \item We analyze in detail the performance of different VLMs that vary in computational complexity and generative capabilities.
    \item We perform a detailed study of digestive tissue type variations by evaluating 3507 digestive WSIs spanning seven different tissue types, which describe the cancer invasiveness and dysplasia status.
    \item We evaluate the models' performance using attention maps at the slide level, conducting a concordance study to assess diagnostically relevant regions, which are validated by a certified pathologist.
\end{enumerate}

% Prior studies have not systematically evaluated prompt design for slide-level diagnostic subtyping. Our work uniquely focuses on dysplasia and invasiveness classification using a structured, multidimensional prompt framework.

\section{Methods}
This section presents the architectures and foundations of three distinct VLM frameworks investigated in this article, namely Quilt-Net, Quilt-LLAVA, and CONCH.

\begin{figure}[ht]
    \centering
    \begin{subfigure}[b]{0.45\textwidth}
        \centering
        \includegraphics[width=\textwidth]{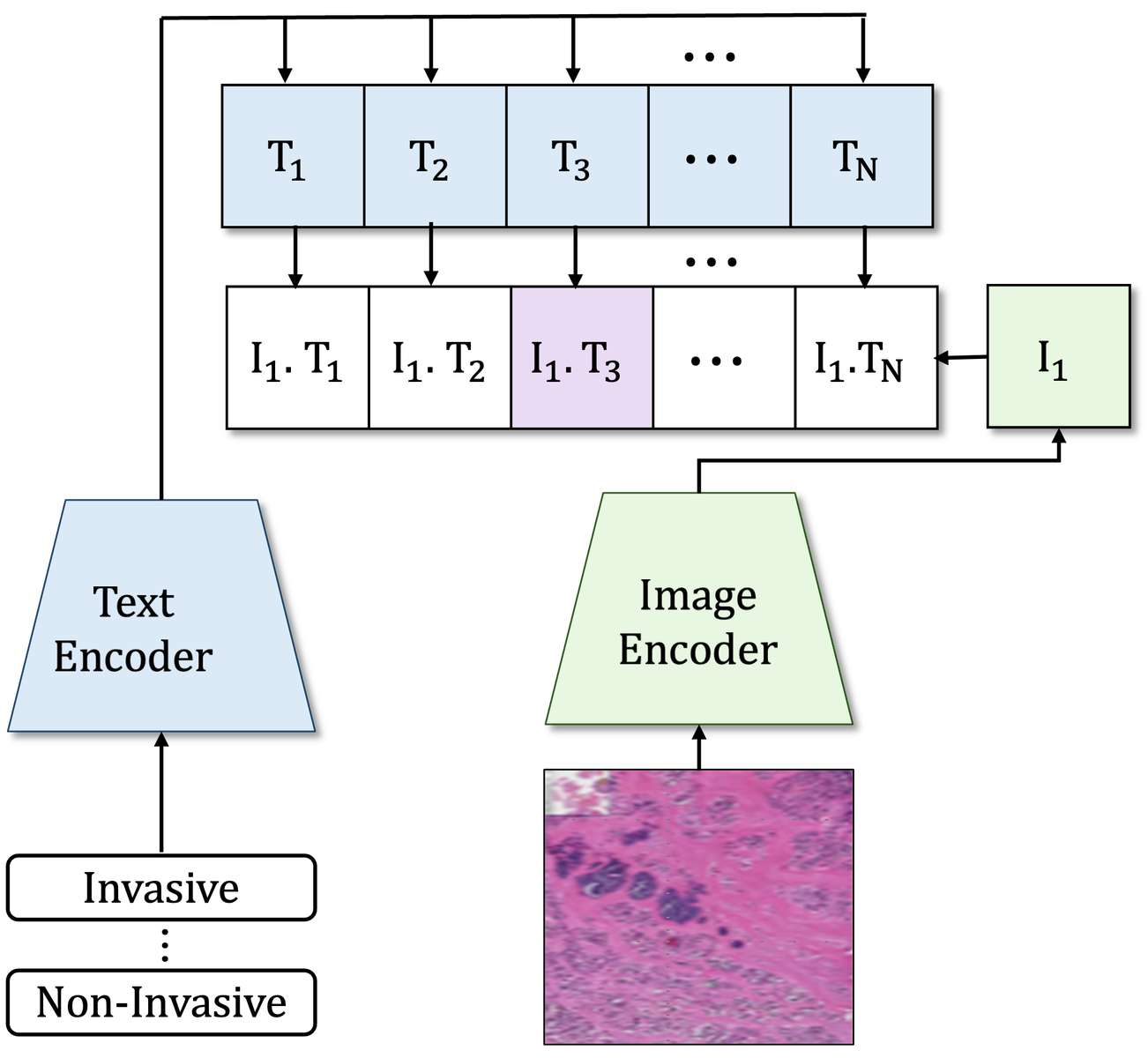}
        \caption{Quilt-Net}
        \label{fig:model1}
    \end{subfigure}
    \hspace{2em}
    \begin{subfigure}[b]{0.27\textwidth}
        \centering
        \includegraphics[width=\textwidth]{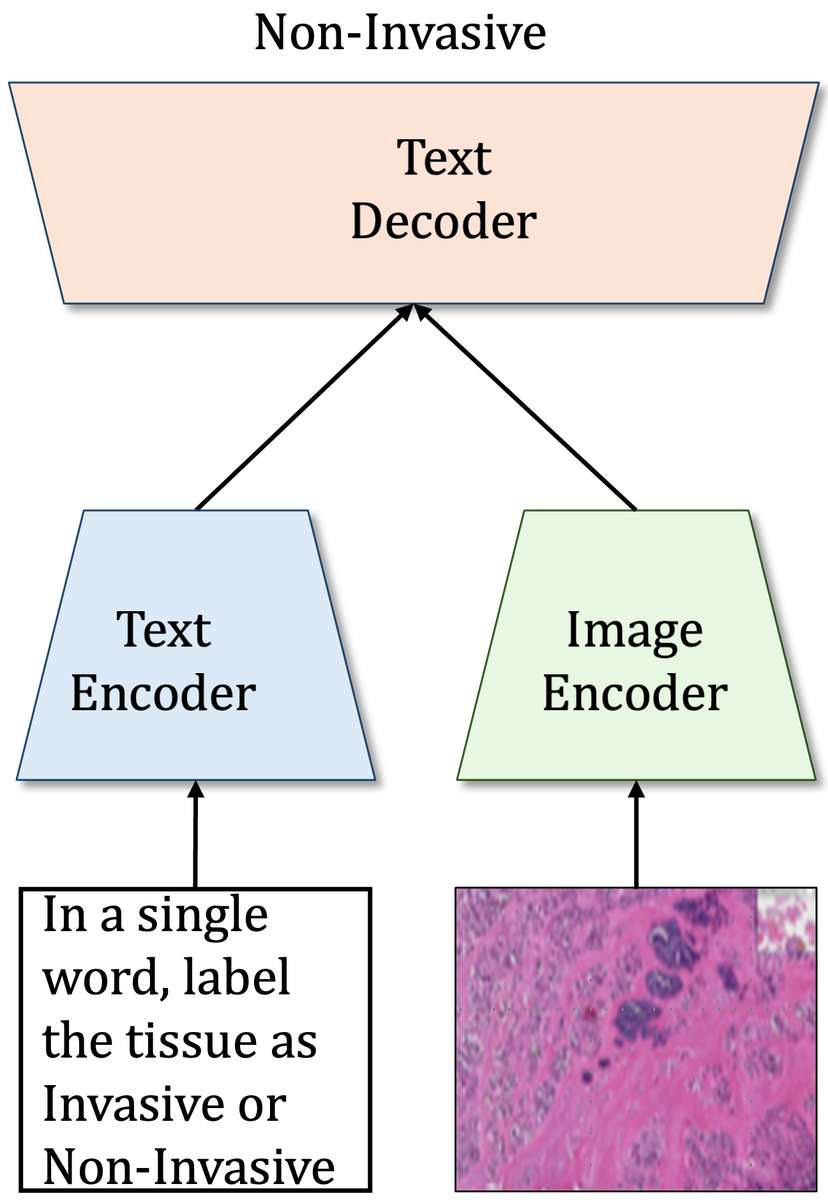}
        \caption{Quilt-LLAVA / CONCH}
        \label{fig:model2}
    \end{subfigure}
    \caption{High level overview of the inference process for the three VLMs.}
    \label{fig:models}
\end{figure}

Quilt-Net establishes a foundational framework for learning robust visual representations from WSIs while simultaneously aligning these representations with natural language descriptions \cite{ikezogwo2025quilt1mmillionimagetextpairs, lu2023visuallanguagepretrainedmultiple}. Drawing inspiration from CLIP's dual-encoder paradigm \cite{radford2021learningtransferablevisualmodels}, Quilt-Net combines the strengths of both contrastive learning and hierarchical feature extraction to address the unique challenges of computational pathology. This approach builds upon recent advancements in self-supervised learning for gigapixel histopathology images \cite{chen2022scalingvisiontransformersgigapixel} and demonstrates superior performance compared to models pre-trained on natural images \cite{5206848}. For a batch of $N$ (image, text) pairs, CLIP attempts to optimize a contrastive objective to create a joint embedding space between image and text embeddings. During inference for a classification task, as summarized in Fig. \ref{fig:model1}, an input image is fed to the image encoder and the class labels are fed to the text encoder. The image and text embeddings then undergo cosine similarity, where the image-text combination with the highest similarity is selected as the class label. The image encoder is based on the ViT-B/32 architecture \cite{dosovitskiy2021imageworth16x16words}, while the text encoder is based on GPT-2 \cite{radford2019language}. Quilt-Net is trained by finetuning the pre-trained CLIP model from OpenAI \cite{radford2021learningtransferablevisualmodels} on Quilt-1M; a diverse dataset of 1M image-text samples. 

Quilt-LLAVA extends beyond the dual-encoder approach of Quilt-Net by adopting the LLAVA framework \cite{liu2023visualinstructiontuning}, which integrates a large language model (LLM) for enhanced vision-language capabilities in computational pathology. This architectural approach enables sophisticated interaction between visual histopathological data and medical textual descriptions, addressing limitations in previous models that lacked generative capabilities \cite{li2023llavamedtraininglargelanguageandvision, sun2024drllavavisualinstructiontuning}. In the Quilt-LLAVA architecture, generally described in Fig. \ref{fig:model2}, the input image goes through a visual encoder (i.e. pre-trained Quilt-Net \cite{ikezogwo2025quilt1mmillionimagetextpairs}) to extract features that are then projected into embeddings. On the other hand, the text input undergoes tokenization and embedding. The image and text embeddings are then concatenated and fed to the LLM for processing. Quilt-LLAVA is trained on Quilt-Instruct; a large-scale dataset of $\sim$ 107k histopathology question/answer pairs associated with images.  

CONCH builds upon the foundations of Quilt-Net and Quilt-LLAVA while introducing novel components for contextual reasoning and knowledge integration \cite{chen2023generalpurposeselfsupervisedmodelcomputational}. Drawing inspiration from the CoCa method and recent advances in VLMs \cite{yu2022cocacontrastivecaptionersimagetext}, CONCH employs a decoupled decoder design that simultaneously supports contrastive and generative objectives. CONCH consists of an image encoder, a text encoder, and a multi-modal text decoder. The image encoder is based on ViT-base and is responsible for transforming the input image into image tokens through a series of transformer and attention pooling layers. The text encoder and multimodal decoder are both GPT-style models. The text encoder processes the textual input into tokens, which are then fused with the image tokens at the multi-modal text decoder. From a high-level view, CONCH has a similar inference pipeline as Quilt-LLAVA, as shown in Fig. \ref{fig:model2}. CONCH is trained on over 1.17M image-caption pairs from histopathology images and biomedical text.

Quilt-Net, Quilt-LLAVA, and CONCH represent distinct vision-language architectures for computational pathology, each with different parameter scales and architectural approaches \cite{lu2023visuallanguagefoundationmodelcomputational, chanda2024neweracomputationalpathology}. Quilt-Net employs a CLIP-inspired dual-encoder approach with approximately 150M parameters (86M for the ViT-B/32 image encoder and 64M for the text encoder), establishing effective contrastive learning between histopathological images and text. Quilt-LLAVA significantly expands this capacity by integrating a large language model with the visual encoder, increasing the parameter count to approximately 7B parameters, enabling more sophisticated reasoning while maintaining a lightweight projection layer. CONCH, with approximately 200M parameters (110M for the language model and 90M for the ViT-B/16 vision encoder), introduces a CoCa-inspired decoupled decoder architecture that efficiently supports both contrastive objectives through a unified framework. This design offers computational advantages while still outperforming general-purpose VLMs on histopathology tasks, with performance gains of 15-20\% on cancer subtyping and prognostic prediction compared to models without hierarchical visual processing capabilities \citep{chen2022scalingvisiontransformersgigapixel, vorontsov2024virchowmillionslidedigitalpathology}.

\section{Benchmarking on Big-Data Cohort of Digestive Pathology}
\begin{figure}[ht]
    \centering
    \includegraphics[width=1\linewidth]{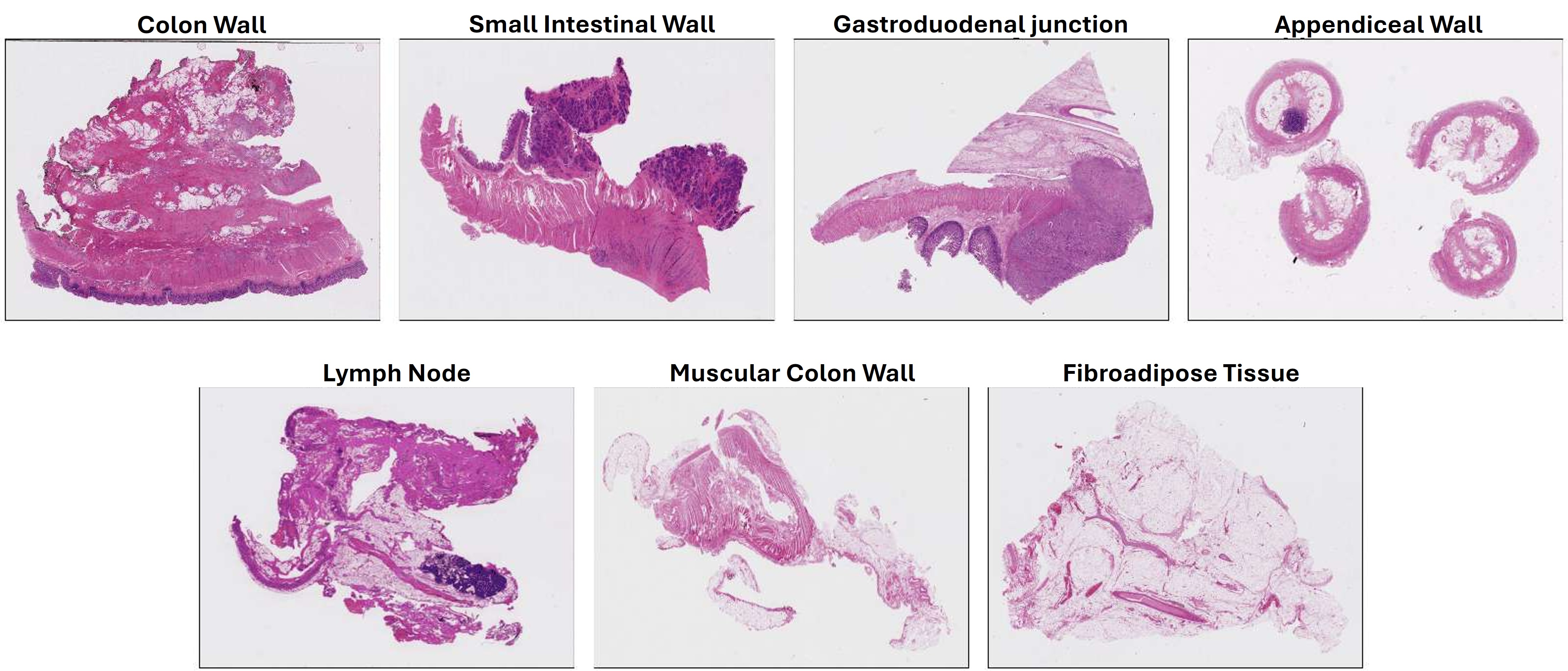}
    \caption{Sample images from the in-house dataset}
    \label{data samples}
\end{figure}

This study leverages a digestive computational pathology dataset obtained through secondary use of giga-pixel WSIs generated during routine clinical care at the Centre Hospitalier de l'Université de Montréal (CHUM) in QC, Canada, with ethics approval 2024-11800, 23.189 - APR. The dataset comprises a comprehensive collection of 3,507 high-resolution WSIs  in big-data form encompassing diverse tissue specimens from the digestive system. Each WSI is annotated on the slide level, providing rich material for our vision-language modeling experiments. The dataset includes seven distinct tissue types with varying representation across classes. Colon wall (CW) specimens constitute the largest proportion (36.18\%, $n=1,269$), followed by lymph nodes (LN) (28.40\%, $n=996$), fibroadipose tissue (FT) (17.65\%, $n=619$), and small intestinal wall (SIW)(12.26\%, $n=430$). The remaining specimens include appendiceal wall (AW) (3.08\%, $n=108$), muscular colon wall (MCW) (1.28\%, $n=45$), and gastroduodenal junctions (GJ) between the colon and small intestine (1.14\%, $n=40$). Each WSI in the dataset is annotated in terms of the dysplasia status and presence of invasive cancer. Table \ref{tab:combined_distribution} summarizes the dataset statistics, while Fig. \ref{data samples} presents sample WSI thumbnails from each tissue type.

\begin{table}[h!]
\centering
\caption{In-house digestive dataset statistics}
\label{tab:combined_distribution}
\begin{tabular}{|l|cc|ccc|cc|}
\hline
\multirow{2}{*}{\textbf{Tissue Type}} & \multicolumn{2}{c|}{\textbf{Distribution}} & \multicolumn{3}{c|}{\textbf{Dysplasia Status}} & \multicolumn{2}{c|}{\textbf{Invasiveness}} \\
\cline{2-8}
 & \textbf{Count} & \textbf{\%} & \textbf{None} & \textbf{Low} & \textbf{High} & \textbf{Non-inv.} & \textbf{Inv.} \\
\hline
Colon wall & 1,269 & 36.18\% & 1,038 & 65 & 166 & 809 & 460 \\
Lymph node & 996 & 28.40\% & 996 & 0 & 0 & 845 & 151 \\
Fibroadipose tissue & 619 & 17.65\% & 619 & 0 & 0 & 521 & 98 \\
Small intestinal wall & 430 & 12.26\% & 413 & 7 & 10 & 364 & 66 \\
Appendiceal wall & 108 & 3.08\% & 106 & 0 & 2 & 104 & 4 \\
Muscular colon wall & 45 & 1.28\% & 44 & 0 & 1 & 29 & 16 \\
Gastroduodenal junction & 40 & 1.14\% & 36 & 1 & 3 & 21 & 19 \\
\hline
\textbf{Total} & \textbf{3,507} & \textbf{100\%} & \textbf{3,252} & \textbf{73} & \textbf{182} & \textbf{2,693} & \textbf{814} \\
\textbf{Percentage} & - & - & \textbf{92.73\%} & \textbf{2.08\%} & \textbf{5.19\%} & \textbf{76.79\%} & \textbf{23.21\%} \\
\hline
\end{tabular}
\end{table}

Dysplasia is the abnormal growth or development of cells, tissues, or organs, typically characterized by altered size, shape, and organization. The majority of specimens in the dataset show no dysplasia (92.73\%, $n=3,252$). High-grade dysplasia is present in 5.19\% ($n=182$) of specimens, while low-grade dysplasia is detected in 2.08\% ($n=73$). This imbalance reflects real world clinical scenarios where pathological findings often represent a small subset of examined tissue. Importantly, the distribution of dysplasia varies considerably across tissue types. Dysplastic changes are predominantly observed in CW specimens, where 18.20\%  of CW specimens exhibit some degree of dysplasia (13.08\% high-grade and 5.12\% low-grade). Dysplasia is also present to a lesser extent in SIW (3.95\%) and GJ specimens (10.00\%), while being rare or absent in LN and FT.

Invasiveness refers to the ability of abnormal cells, particularly cancer cells, to penetrate and infiltrate surrounding tissues, breaking through basement membranes and potentially spreading to distant sites. Approximately one-quarter of all specimens (23.21\%, $n=814$) exhibit invasive characteristics, while the majority (76.79\%, $n=2,693$) are non-invasive. The distribution of invasiveness varies significantly across tissue types, revealing valuable patterns for model learning. The highest rates of invasion are observed in gastroduodenal junction specimens (47.50\%), followed by colon wall (36.25\%), muscular colon wall (35.56\%), and lymph nodes (15.16\%). This event is a typical finding during routine pathology diagnostics and is critical to clinical care as the degree of invasion into tissue layers and types are the most significant predictor of cancer aggressiveness. It presents a clinically relevant and critical task that can be addressed by vision-language models to learn contextually relevant associations. 

\section{Evaluation Methodology}\label{Sec: Eval Meth}
This section discusses the methodology followed to assess the aforementioned VLMs, i.e. Quilt-Net, Quilt-LLAVA, and CONCH on the in-house digestive dataset.  All models were initialized with pre-trained weights from their respective base architectures without further fine-tuning. We investigate how prompt engineering affects the models' ability to identify invasiveness and dysplasia status across diverse digestive system tissue samples. We processed WSIs from our dataset by extracting patches at 5$\times$ magnification level using a sliding window approach with a patch size of 512×512 pixels and 0\% overlap. We developed a systematic prompt engineering framework based on information theory and clinical communication principles to evaluate how linguistic variations influence model performance in computational pathology tasks.

\begin{table}[h!]
\small
\caption{Prompt templates for histopathology Invasive classification}
\label{tab:prompts}
\begin{tabular}{|p{0.9cm}|p{2.7cm}|p{8cm}|}
\hline
\textbf{ID} & \textbf{Dimensions} & \textbf{Template} \\
\hline
$P_1(O)$ & \makecell[l]{\textbf{\textit{DS}}: \text{Medium}\\ \textbf{\textit{AP}}: \text{High}\\ \textbf{\textit{IF}}: \text{Minimal}\\ \textbf{\textit{OC}}: \text{Explicit}} & 
 \makecell[l]{``The image is taken from the $O$ using H\&E staining,\\ output only the label name which best fits the image\\ out of the following Invasive or Non-Invasive''} \\
\hline
$P_2(O)$ & \makecell[l]{\textbf{\textit{DS}}: \text{Medium}\\ \textbf{\textit{AP}}: \text{High}\\ \textbf{\textit{IF}}: \text{Minimal}\\ \textbf{\textit{OC}}: \text{Explicit}} & 
\makecell[l]{``The image is taken from the $O$ using H\&E staining,\\ output only the label name which best fits the image\\ out of the following Cancerous or Normal''} \\
\hline
$P_3(O)$ & \makecell[l]{\textbf{\textit{DS}}: \text{High}\\ \textbf{\textit{AP}}: \text{High}\\ \textbf{\textit{IF}}: \text{Expert}\\ \textbf{\textit{OC}}: \text{Explicit}} & 
\makecell[l]{``You are an expert pathologist analyzing histopathology\\ slides. Given an image of a tissue sample stained with\\ Hematoxylin and Eosin (H\&E) from the $O$ and the\\ question of classifying the presence of cancer, classify\\ it as either `Invasive' or `Non-Invasive'. Provide only\\ the single word label.''} \\
\hline
$P_4(O)$ & \makecell[l]{\textbf{\textit{DS}}: \text{High}\\ \textbf{\textit{AP}}: \text{High}\\ \textbf{\textit{IF}}: \text{Task}\\ \textbf{\textit{OC}}: \text{Explicit}} & 
\makecell[l]{``Given an image of a tissue sample stained with\\ Hematoxylin and Eosin (H\&E) from the $O$, classify\\ the existence of cancer as either `Invasive' or\\ `Non-Invasive'. Provide only a single word label.''} \\
\hline
$P_5(O)$ & \makecell[l]{\textbf{\textit{DS}}: \text{Medium}\\ \textbf{\textit{AP}}: \text{High}\\ \textbf{\textit{IF}}: \text{Task}\\ \textbf{\textit{OC}}: \text{Explicit}} & 
\makecell[l]{``Given an image of a tissue sample stained with\\ hematoxylin and eosin from the $O$, identify whether the\\ sample is cancerous or not. Provide only a single word\\ label''} \\
\hline
$P_6$ & \makecell[l]{\textbf{\textit{DS}}: \text{Medium}\\ \textbf{\textit{AP}}: \text{Medium}\\ \textbf{\textit{IF}}: \text{Task}\\ \textbf{\textit{OC}}: \text{Explicit}} & 
\makecell[l]{``Given an image of a tissue sample stained with\\ hematoxylin and eosin from the gastrointestinal system,\\ identify whether the sample is cancerous or not.\\ Provide only a single word label''} \\
\hline
$P_7$ & \makecell[l]{\textbf{\textit{DS}}: \text{Medium}\\ \textbf{\textit{AP}}: \text{Medium}\\ \textbf{\textit{IF}}: \text{Task}\\ \textbf{\textit{OC}}: \text{Explicit}} & 
\makecell[l]{``Given an image of a tissue sample stained with\\ hematoxylin and eosin from the digestive system,\\ identify whether the sample is cancerous or not. Provide\\ only a single word label''} \\
\hline
$P_8$ & \makecell[l]{\textbf{\textit{DS}}: \text{Medium}\\ \textbf{\textit{AP}}: \text{Low}\\ \textbf{\textit{IF}}: \text{Task}\\ \textbf{\textit{OC}}: \text{Explicit}} & 
\makecell[l]{``Given an image of a tissue sample stained with\\ hematoxylin and eosin, identify whether the sample\\ is cancerous or not. Provide only a single word label''} \\
\hline
$P_9$ & \makecell[l]{\textbf{\textit{DS}}: \text{High}\\ \textbf{\textit{AP}}: \text{Medium}\\ \textbf{\textit{IF}}: \text{Expert}\\ \textbf{\textit{OC}}: \text{Explicit}} & 
\makecell[l]{``As a pathologist examining this H\&E-stained\\ digestive system tissue sample, provide your\\ assessment of malignancy as a single word: either\\ `Invasive' or `Non-Invasive'.''} \\
\hline
\end{tabular}
\end{table}

Our framework explores four critical dimensions of prompt design that we hypothesize significantly impact model performance. We formalized a set of nine prompt templates by systematically varying four key dimensions: detail specificity ($DS$), anatomical precision ($AP$), instructional framing ($IF$), and output constraints ($OC$). In templates where organ-specific information is required, we use the variable $O$ as a placeholder for the target organ being examined. $DS$ refers to the level of granularity in the prompt, ranging from general to detailed instructions, taking values of LOW, MEDIUM, or HIGH. $AP$ represents the extent to which the prompt includes precise anatomical details to make the prompt more focused, similarly taking values of LOW, MEDIUM, or HIGH. $IF$ determines the structure of the prompt, such as posing a direct question versus providing a declarative statement, with values of EXPERT (positioning the model as a specialist), MINIMAL (providing basic instructions), or TASK (focusing on specific objectives). $OC$ controls the format and length of the model's response to ensure consistency, defined as either EXPLICIT (strictly defined output format) or IMPLICIT (loosely defined format). Table~\ref{tab:prompts} presents each prompt template with its corresponding dimensional properties.

The prompt templates were strategically designed to address several research questions in medical vision-language interaction:

\begin{enumerate}
    \item \textbf{Information Theoretic Perspective}: We hypothesized that intermediate levels of information content in prompts (neither too sparse nor too detailed) would optimize model performance, following principles from communication theory and cognitive load theory \cite{Wang_2024}. Prompts 3-5 were designed with varying information density to test this hypothesis.

    \item \textbf{Anatomical Specificity Gradient}: Prompts 5-8 implement a controlled degradation of anatomical specificity to quantify how precision of anatomical reference affects classification performance. This addresses a key question in medical AI regarding the importance of anatomical context in diagnostic reasoning.

    \item \textbf{Expert Role Framing}: Prompts 3 and 9 incorporate expert role assignment, a technique that has shown promise in general LLM task performance but remains under-explored in medical vision-language tasks. By positioning the model as a pathologist, we investigated whether role framing enhances performance on specialized medical tasks.

    \item \textbf{Output Constraint Consistency}: All prompts maintain explicit output constraints to isolate the effects of input prompt variations rather than confounding with output format variations.
\end{enumerate}

This systematic approach to prompt design allows us to quantify the relationship between linguistic features of prompts and model performance, potentially yielding insights for optimal prompt engineering in medical vision-language applications.

\section{Results}
This section summarizes the performance of the different VLM models under various prompts on our in-house digestive dataset. We first conduct an ablative study comparing the different performances using Receiver Operating Characteristic (ROC) curves and Area Under the Curve (AUC) scores. We mainly analyze the effect of prompt design and model complexity on the performance of cancer invasiveness classification. We then use the task of dysplasia status classification as a means to conduct focused analysis on the best prompt. We also analyze attention maps obtained by the different models on different WSIs and highlight relevant tissue regions, with feedback given by a certified pathologist. All experiments were conducted on NVIDIA A100-SXM4-40GB GPUs to ensure consistent evaluation across all models. Quilt-LLAVA during inference uses a temperature of 0.1 for minimized hallucinations and less variability in output.

\subsection{Ablative study}
\begin{figure}[ht!]
    \centering
    \begin{subfigure}[b]{0.32\textwidth}
        \centering
        \includegraphics[width=\textwidth]{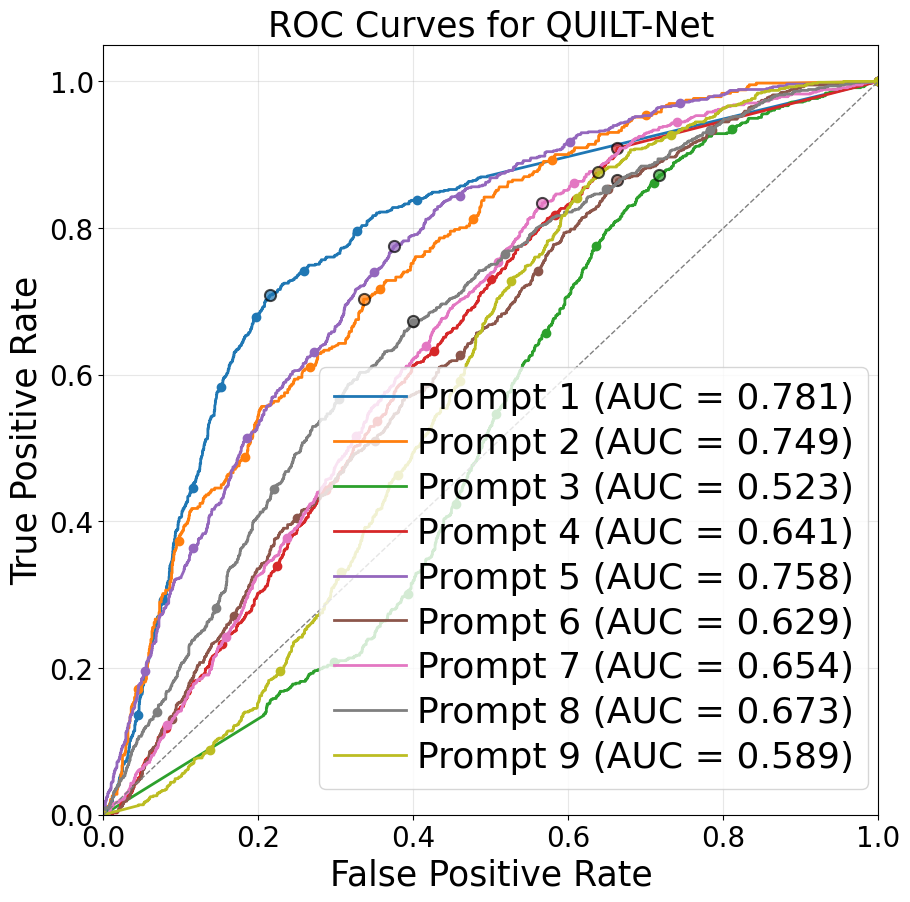}
        \caption{Quilt-Net}
        \label{fig:roc_quiltNet}
    \end{subfigure}
    \hfill
    \begin{subfigure}[b]{0.32\textwidth}
        \centering
        \includegraphics[width=\textwidth]{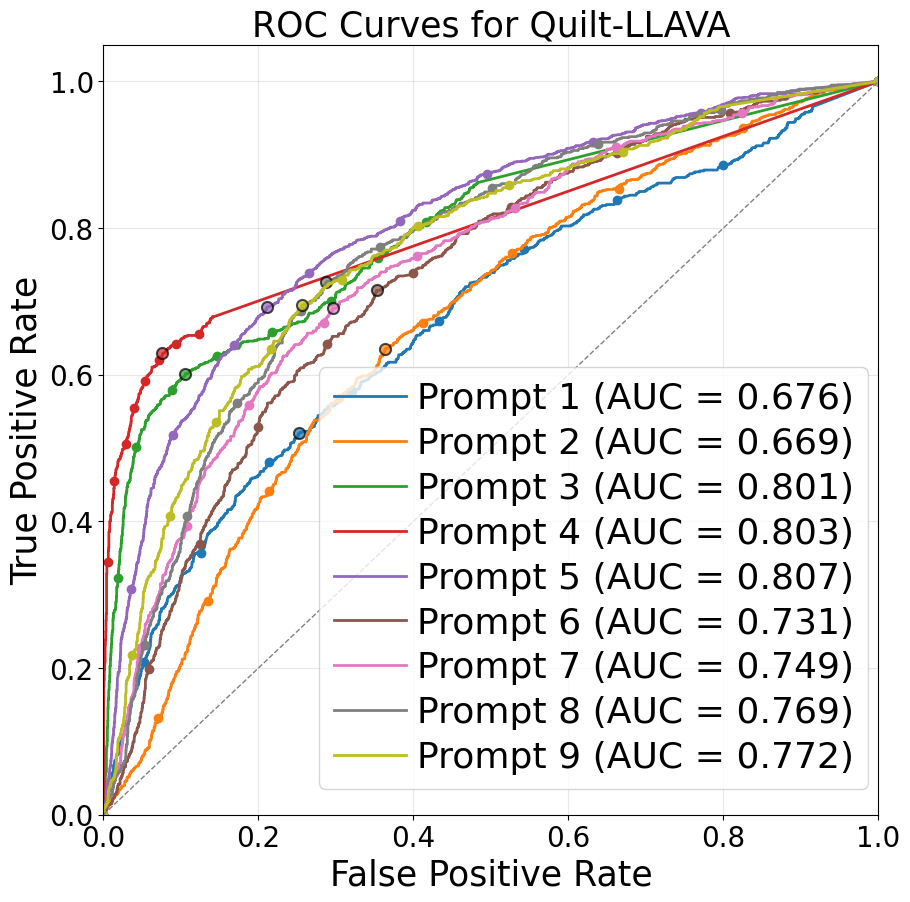}
        \caption{Quilt-LLAVA}
        \label{fig:roc_quilt}
    \end{subfigure}
    \hfill
    \begin{subfigure}[b]{0.32\textwidth}
        \centering
        \includegraphics[width=\textwidth]{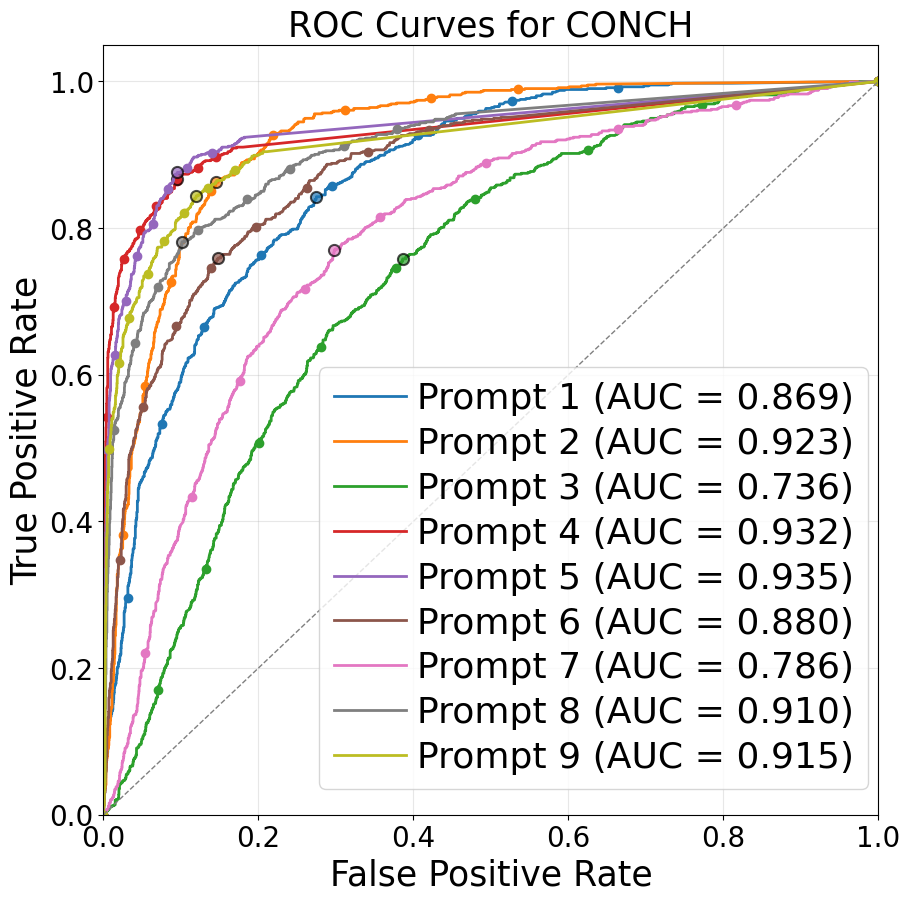}
        \caption{CONCH}
        \label{fig:roc_conch}
    \end{subfigure}
    % \hfill
    % \begin{subfigure}[b]{0.24\textwidth}
    %     \centering
    %     \includegraphics[width=\textwidth]{Figures/Image7.png}
    %     \caption{BioMedCLIP}
    %     \label{fig:roc_biomed}
    % \end{subfigure}

    \caption{ROC curves comparing the performance of three vision-language models: (a) Quilt-Net, (b) Quilt-LLAVA, and (c) CONCH, across different prompts for invasive cancer classification. CONCH demonstrates the highest robustness and performance consistency, while Quilt-Net and Quilt-LLAVA exhibit significant sensitivity to prompt design.}
    \label{fig:roc_comparison}
\end{figure}

\begin{figure}[ht!]
    \centering
    \begin{subfigure}[b]{0.50\textwidth}
        \centering
        \includegraphics[width=\textwidth]{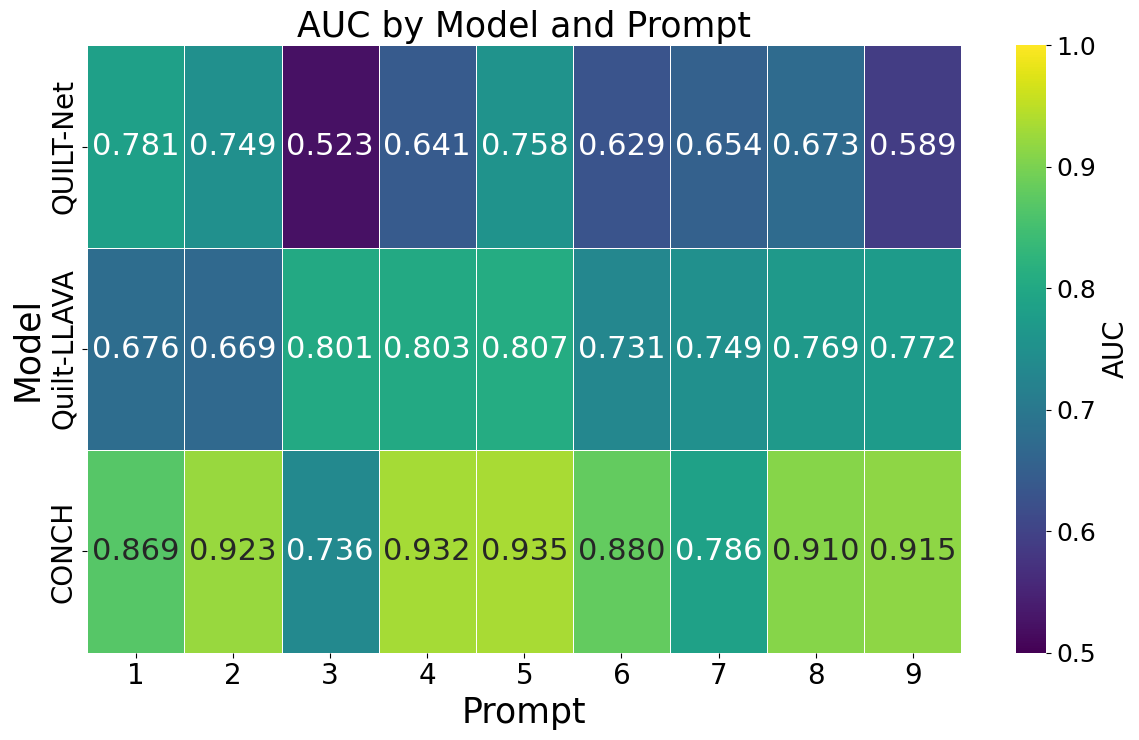}
        \caption{AUC heatmap by model and prompt.}
        \label{fig:avg_roc_heatmap}
    \end{subfigure}
    \hspace{1em}
    \begin{subfigure}[b]{0.43\textwidth}
        \centering
        \includegraphics[width=\textwidth]{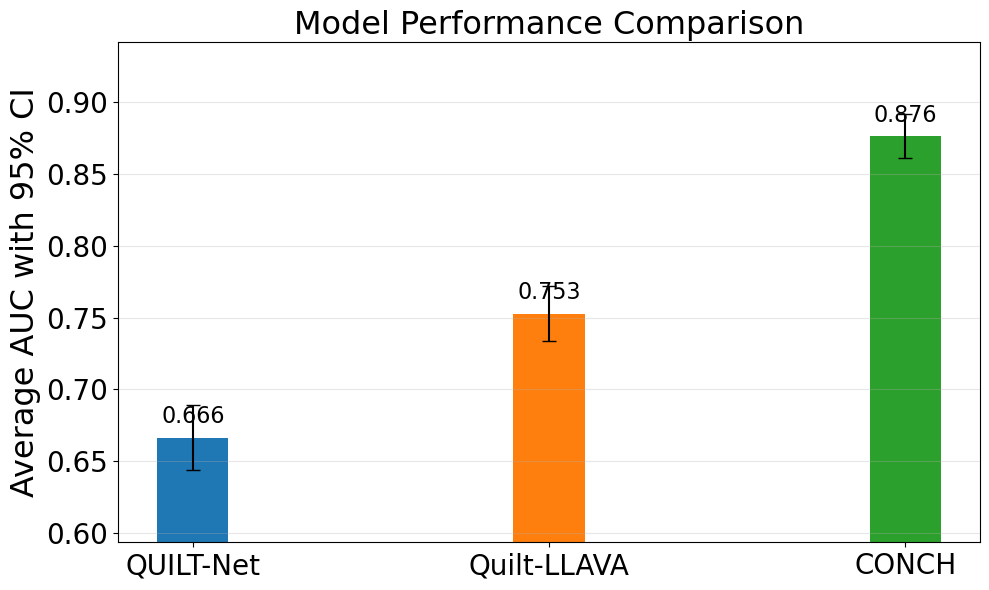}
        \caption{Average AUC performance comparison.}
        \label{fig:model_comparison}
    \end{subfigure}
    \caption{Performance comparison of VLM models in terms of ROC AUC. CONCH consistently outperforms the other models despite having fewer parameters than Quilt-LLAVA, underscoring the importance of domain-specific training and prompt alignment over model scale.}
    \label{fig:performance_comparison}
\end{figure}

Our ablative study starts by investigating the impact of prompt formulations on model performance. Figs. \ref{fig:roc_comparison} and \ref{fig:performance_comparison} summarize the ROC curves and AUC results for different prompts used on Quilt-Net, Quilt-LLAVA, and CONCH, demonstrating significant performance variations based on architectural differences and prompt design choices. We evaluated model performance using ROC curves and the corresponding AUC metric, which plot true positive rates against false positive rates at various classification thresholds. The AUC metric ranges from 0 to 1, with higher values indicating superior discriminative ability.

When analyzing the  models' performances, it can be seen that Quilt-Net and Quilt-LLAVA generally show pronounced drops in AUC scores with certain prompts. This is unlike CONCH, which is more robust to most of the changes. When comparing prompts 3-5, which vary in terms of information density, it is evident that more information leads to degraded performance, which is seen in the AUC drop from 0.758 to 0.523 and from 0.935 to 0.736 for Quilt-Net and CONCH, respectively, when going from $P_5$ to $P_3$. On the other hand, Quilt-LLAVA shows consistent performance across the three prompts, indicating that the model is less sensitive to varying information density. This supports our first hypothesis, presented in Section \ref{Sec: Eval Meth}, stating that intermediate levels of general information optimize the model's performance better when compared to those that are too detailed. When comparing prompts 5-8, which vary in terms of anatomical specificity, the importance of precise anatomical context is evident in classification performance. For all models, prompt 5 (which has high anatomical precision) achieves stronger performance (0.758 for Quilt-Net, 0.807 for Quitl-LLAVA, and 0.935 for CONCH) when compared to prompts 6-8 that have lower anatomical specificity. This answers our second research question regarding the importance of anatomical context in diagnostic reasoning. When evaluating the impact of expert role framing (prompts 3 and 9), we observe a negative or neutral effect on performance. Specifically, in Quilt-Net, prompts 3 and 9 show the worst performance, with AUC scores of 0.523 and 0.589, respectively. Similarly, prompt 3 gives the lowest performance in CONCH (AUC=0.736), while prompt 9 shows neutral behavior with no improvement (AUC=0.915). This suggests that framing the model as an expert does not necessarily enhance the model's performance, and may even introduce unnecessary complexity that misleads the model's pre-trained embeddings. On the contrary, Quilt-LLAVA shows good performance for both prompts, which could be attributed to its robust vision-language alignment and effective prompt generalization capabilities.

% \begin{figure}[ht!]
%     \centering
%     \includegraphics[width=0.5\textwidth]{Figures/Image 2.png}
%     \caption{AUC heatmap displaying performance values by model and prompt configuration.}
%     \label{fig:auc_heatmap}
% \end{figure}

Figure \ref{fig:performance_comparison} analyzes the average performance (ROC and AUC) of the three VLMs. As seen in the figure, CONCH achieves the highest average AUC (0.876), followed by Quilt-LLAVA (0.753) and Quilt-Net (0.666). The under-performance by Quilt-Net is expected, given it is the smallest model compared to the other two. However, despite Quilt-LLAVA being the largest model with nearly 7B parameters, it does not outperform CONCH (200M parameters). This suggests that model scale alone is not a dominant factor in performance, and domain-specific training and vision-language alignment have crucial roles. While Quilt-LLAVA uses instruction tuning and a powerful LMM (LLAMA), it is constrained with suboptimal domain alignment between its visual encoder and the LLM for computational pathology. This is unlike CONCH, which uses a contrastive learning approach specifically tuned on histopathology image-text pairs, allowing for better generalizability.

 Table \ref{tab:auc_avg_results} details the aforementioned performances across the different tissue types in our in-house digestive dataset. The results show that GJ and CW achieve the highest classification performance, which is likely due to their distinct histological features and relatively high representation in the dataset (36\% for CW). On the other hand, AW and LN are the most challenging, which is due to their  more complex histopathological variations. Interestingly, FT performed less than expected, despite being relatively well-represented (17.65\%). This could be due to its overlapping visual characteristics with other soft tissues.

\begin{table}[ht!]
\centering
\caption{Average AUC on the different tissue types}
\label{tab:auc_avg_results}
\footnotesize
\setlength{\tabcolsep}{2.8pt} 
\renewcommand{\arraystretch}{1.1} 
\begin{tabular}{|l|*{7}{c|}}
\hline
\textbf{Model} & \multicolumn{7}{c|}{\textbf{Tissue Type}} \\
\cline{2-8}
 & \textbf{CW} & \textbf{SIW} & \textbf{GJ} & \textbf{AW} & \textbf{LN} & \textbf{MCW} & \textbf{FT} \\
\hline
Quilt-Net & 0.76 & 0.68 & 0.88 & 0.55 & 0.61 & 0.75 & 0.75 \\
Quilt-LLAVA & 0.86 & 0.74 & 0.96 & 0.70 & 0.71 & 0.91 & 0.68 \\
CONCH & 0.93 & 0.90 & 0.97 & 0.52 & 0.84 & 0.87 & 0.81 \\
\hline
\textbf{Avg } & \textbf{0.85} & \textbf{0.77} & \textbf{0.94} & \textbf{0.59} & \textbf{0.72} & \textbf{0.84} & \textbf{0.74} \\
\hline
\end{tabular}
\end{table}

\begin{figure}[ht!]
    \centering
    \begin{subfigure}[b]{0.52\textwidth}
        \centering
        \includegraphics[width=\textwidth]{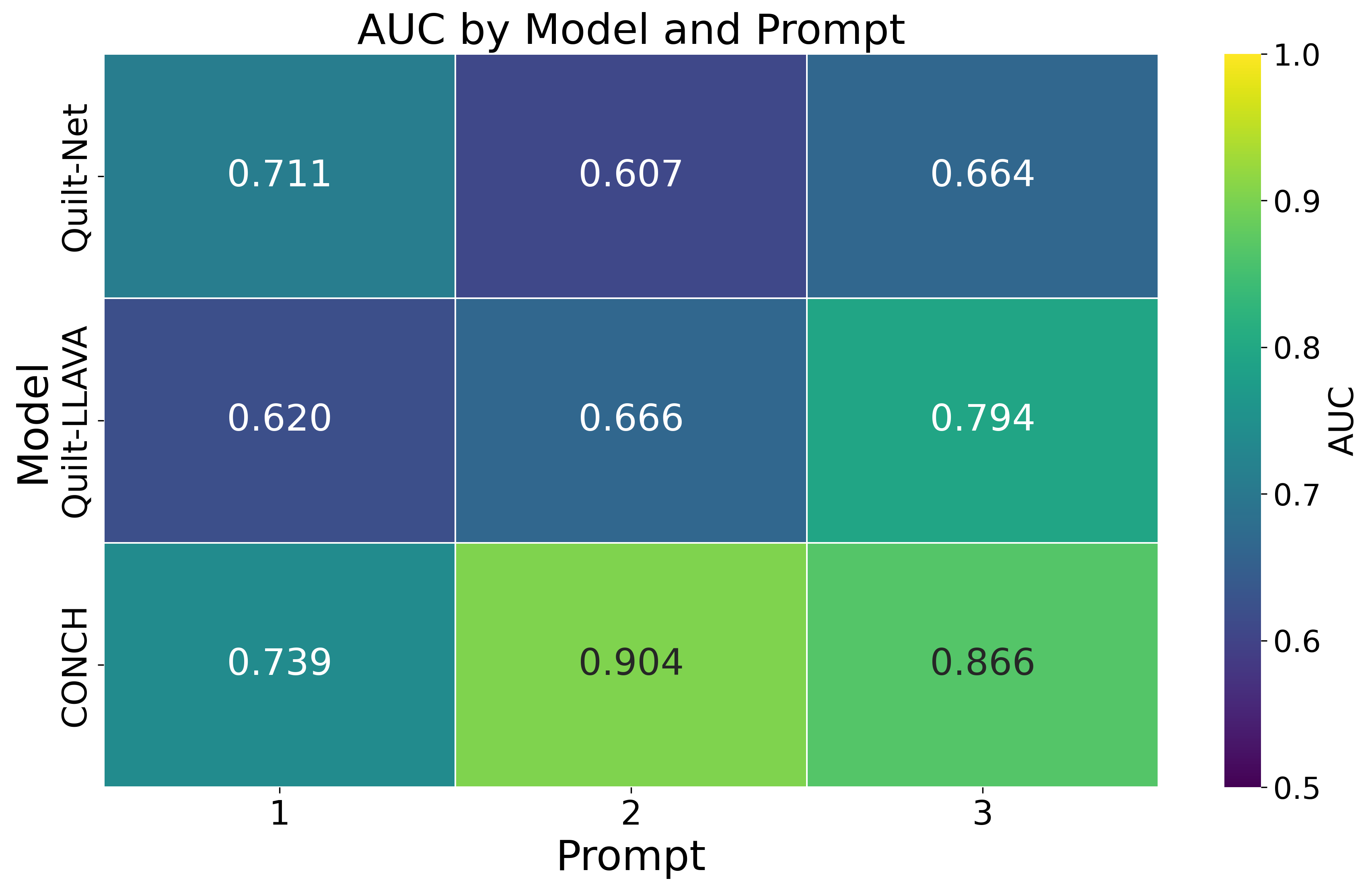}
        \caption{AUC heatmap by model and prompt.}
        \label{fig:avg_roc_heatmap2}
    \end{subfigure}
    \hspace{1em}
    \begin{subfigure}[b]{0.42\textwidth}
        \centering
        \includegraphics[width=\textwidth]{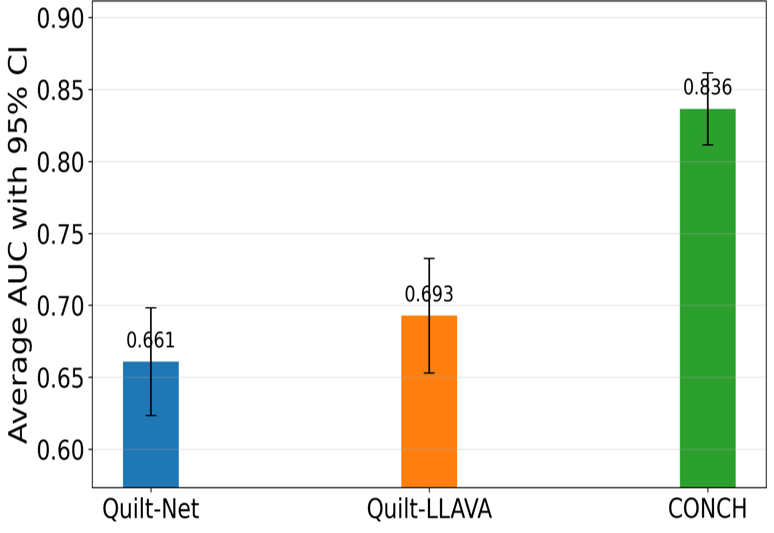}
        \caption{Average AUC performance comparison.}
        \label{fig:model_comparison2}
    \end{subfigure}
    \caption{Performance comparison of VLM models and prompts in terms of ROC AUC for dysplasia status.}
    \label{fig:performance_comparison_dys}
\end{figure}

We extend the analysis to the classification of dysplasia while assessing prompt wording effects on the model performance. We conduct an ablative experiment using three prompt variants derived from the base prompt P$_5$(O). In each variant, the key term for the target pathology was changed while keeping all other prompt aspects constant (medium detail specificity, high anatomical precision, task-oriented instruction, and explicit output constraints). Specifically, D$_1$(O) used the term dysplasia, D$_2$(O) replaced it with atypia, and D$_3$(O) used precancerous. These synonyms describe the same precancerous condition but differ in technical tone as seen in Table \ref{tab:dysplasia_prompts}. The task for each vision-language model remained identifying dysplasia in images given the prompt. Performance was evaluated by AUC, summarized in Fig. \ref{fig:performance_comparison_dys}. As can be seen, CONCH consistently outperforms the other models, emphasizing on its strong capabilities. With regards to the prompt design, it can be seen that the best performing prompt differs from one model to the other. The best performing prompt for Quilt-Net is the one the "Dysplasia" term (AUC=0.711), while the term "Atypia" performed best for Quilt-LLAVA (AUC=0.794) and the term "Precancerous" performed best for CONCH (AUC=0.904). This interesting finding highlights the significant effect of prompt wording, even when using near-synonymous medical terms.

\begin{table}[h!]
\small
\centering
\caption{Prompt templates for histopathology dysplasia classification}
\label{tab:dysplasia_prompts}
\begin{tabular}{|p{0.9cm}|p{2.7cm}|p{8cm}|}
\hline
\textbf{ID} & \textbf{Dimensions} & \textbf{Template} \\
\hline
$D_1(O)$ & \makecell[l]{\textbf{\textit{DS}}: \text{Medium}\\ \textbf{\textit{AP}}: \text{High}\\ \textbf{\textit{IF}}: \text{Task}\\ \textbf{\textit{OC}}: \text{Explicit}} & 
\makecell[l]{``Given an image of a tissue sample stained with\\ hematoxylin and eosin from the $O$, identify whether the\\ sample is Dysplasia or Benign. Provide only \\a single word label''} \\
\hline
$D_2(O)$ & \makecell[l]{\textbf{\textit{DS}}: \text{Medium}\\ \textbf{\textit{AP}}: \text{High}\\ \textbf{\textit{IF}}: \text{Task}\\ \textbf{\textit{OC}}: \text{Explicit}} & 
\makecell[l]{``Given an image of a tissue sample stained with\\ hematoxylin and eosin from the $O$, identify whether the\\ sample is Atypia or Benign. Provide only \\a single word label''} \\
\hline
$D_3(O)$ & \makecell[l]{\textbf{\textit{DS}}: \text{Medium}\\ \textbf{\textit{AP}}: \text{High}\\ \textbf{\textit{IF}}: \text{Task}\\ \textbf{\textit{OC}}: \text{Explicit}} & 
\makecell[l]{``Given an image of a tissue sample stained with\\ hematoxylin and eosin from the $O$, identify whether the\\ sample is Precancerous or Benign. Provide only \\a single word label''} \\
\hline
\end{tabular}
\end{table}

\begin{figure}[ht!]
    \centering
    \includegraphics[width=0.5\textwidth]{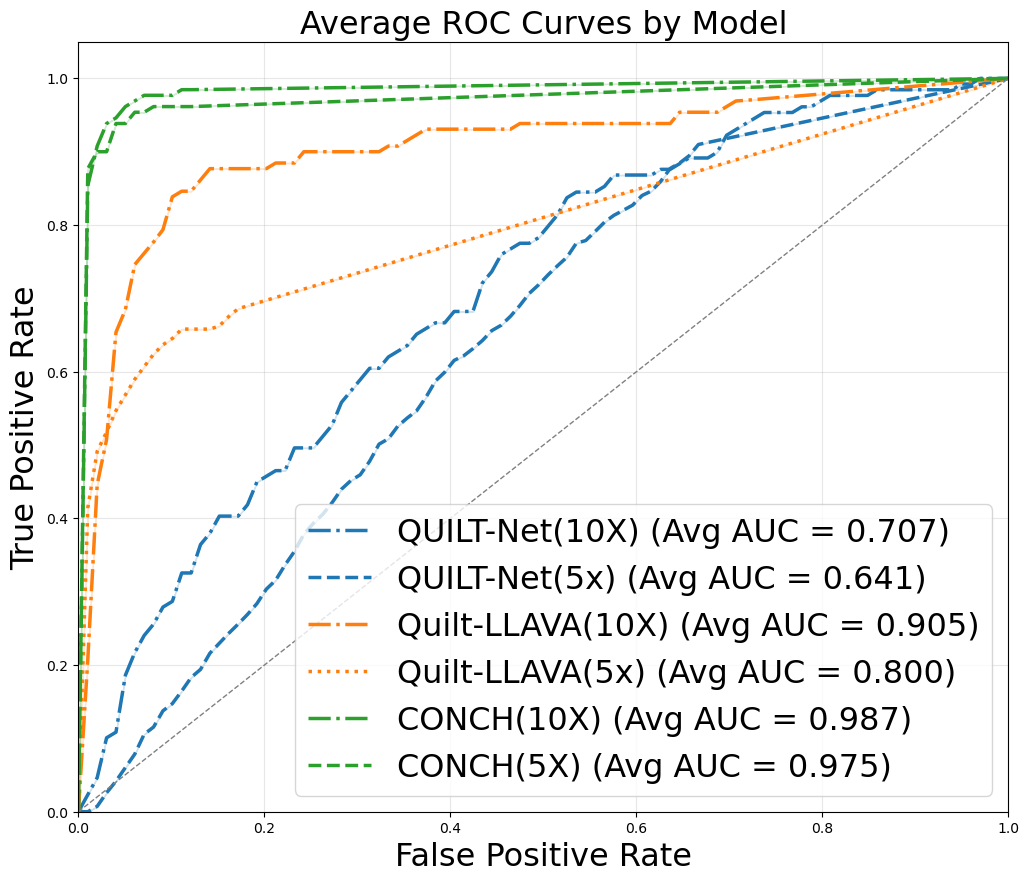}
    \caption{Average ROC curves comparing model performance at different magnification levels.} 
    \label{fig:magnification_comparison}
\end{figure}

Figure \ref{fig:magnification_comparison} analyzes the effect of magnification levels on the model's performance. Generally, it can be seen that better performance is achieved with higher magnification, as it aids in capturing finer morphological structures, which could be needed for accurate classifications. It is also evident that the performance in CONCH is nearly similar regardless of the magnification level, indicating that CONCH is more robust to changes in resolution.

\subsection{Attention Maps Analysis}

\begin{figure}[ht!]
  \centering
  
  % First row: WSI 1
  \begin{subfigure}{\textwidth}
    \centering
    \begin{subfigure}[b]{0.33\textwidth}
      \includegraphics[width=\textwidth]{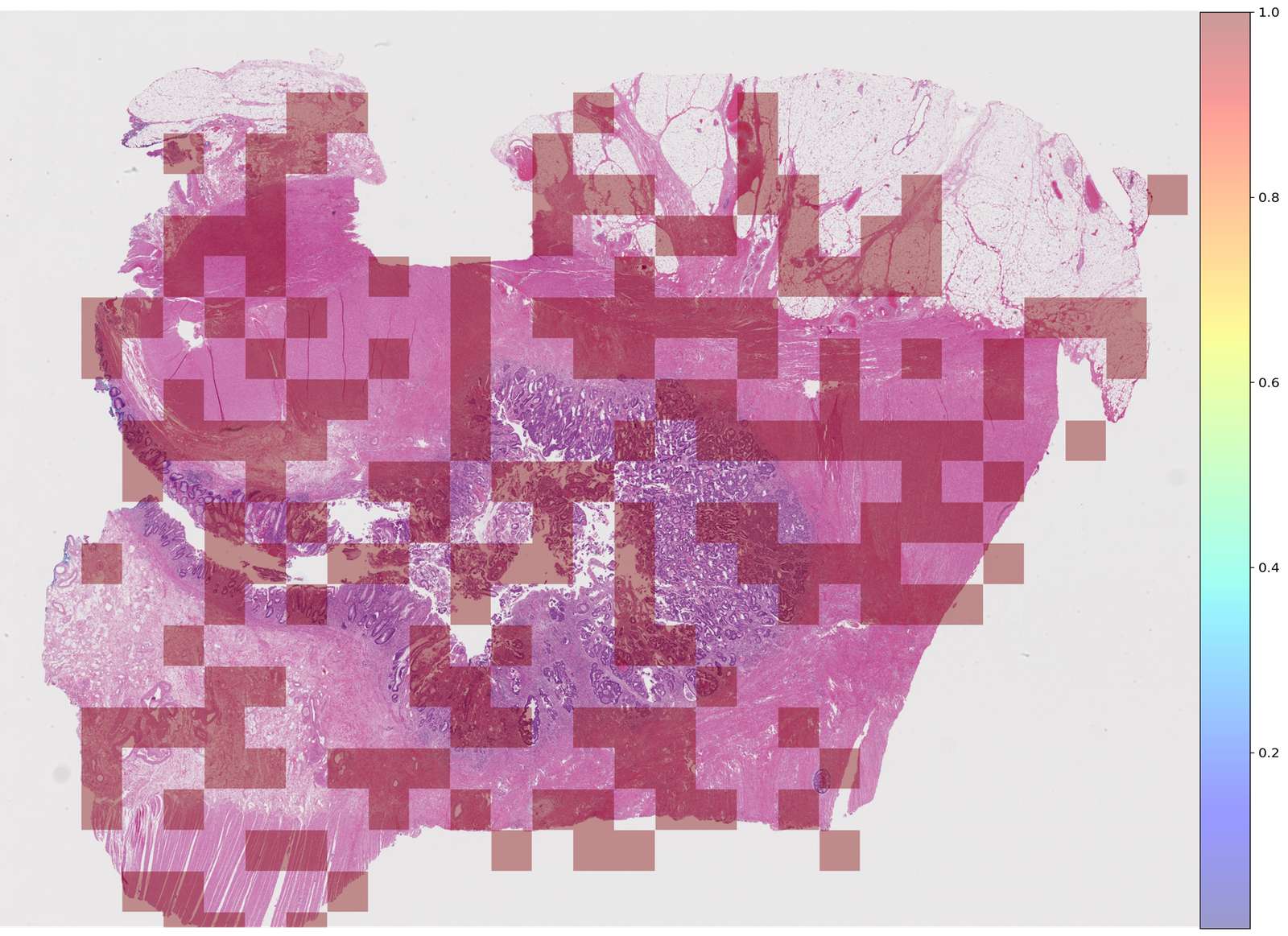}
      \caption{WSI 1: Quilt-Net}
      \label{fig:wsi1-model1}
    \end{subfigure}%
    \hfill
    \begin{subfigure}[b]{0.33\textwidth}
      \includegraphics[width=\textwidth]{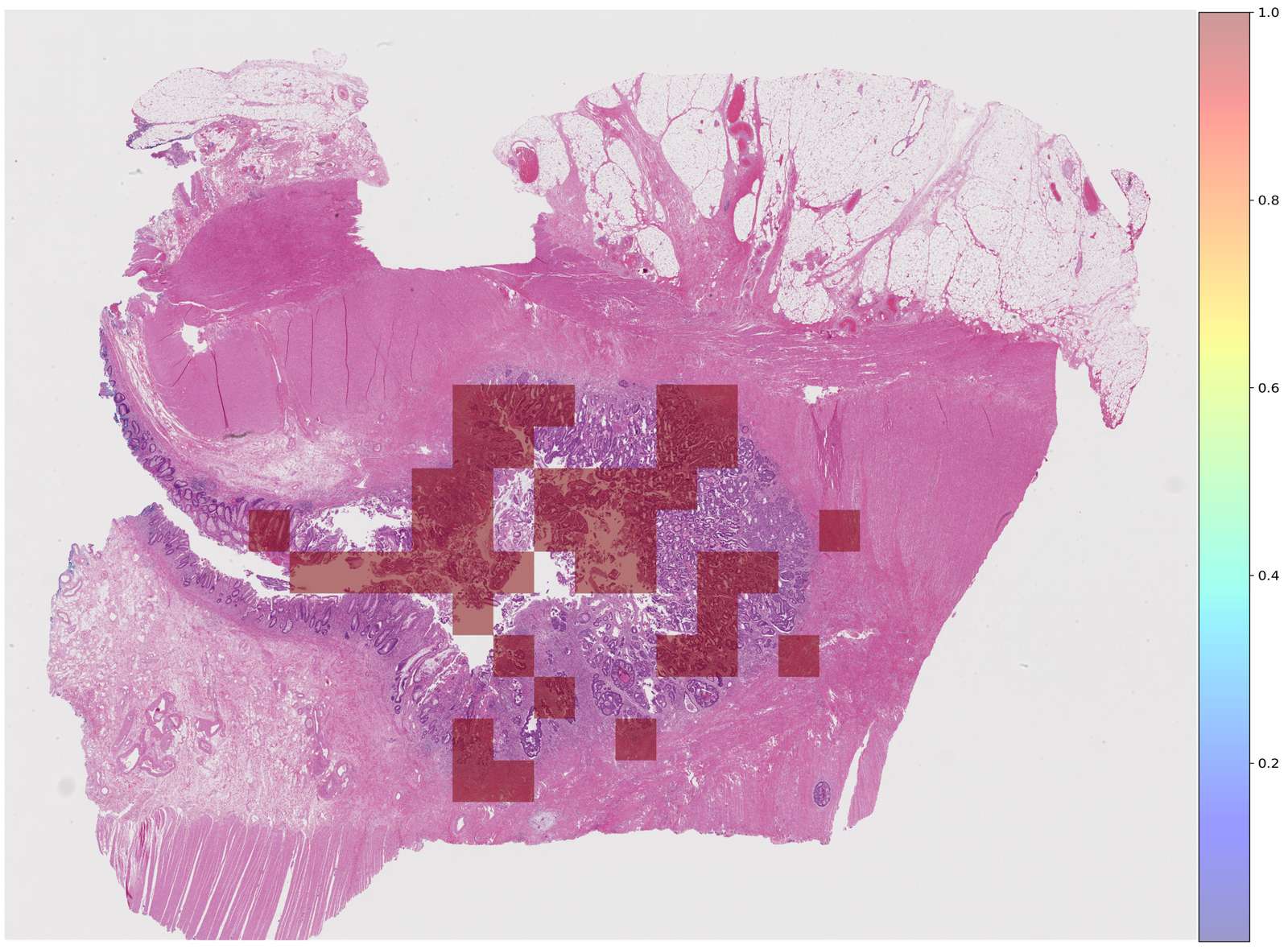}
      \caption{WSI 1: Quilt-LLAVA}
      \label{fig:wsi1-model2}
    \end{subfigure}%
    \hfill
    \begin{subfigure}[b]{0.33\textwidth}
      \includegraphics[width=\textwidth]{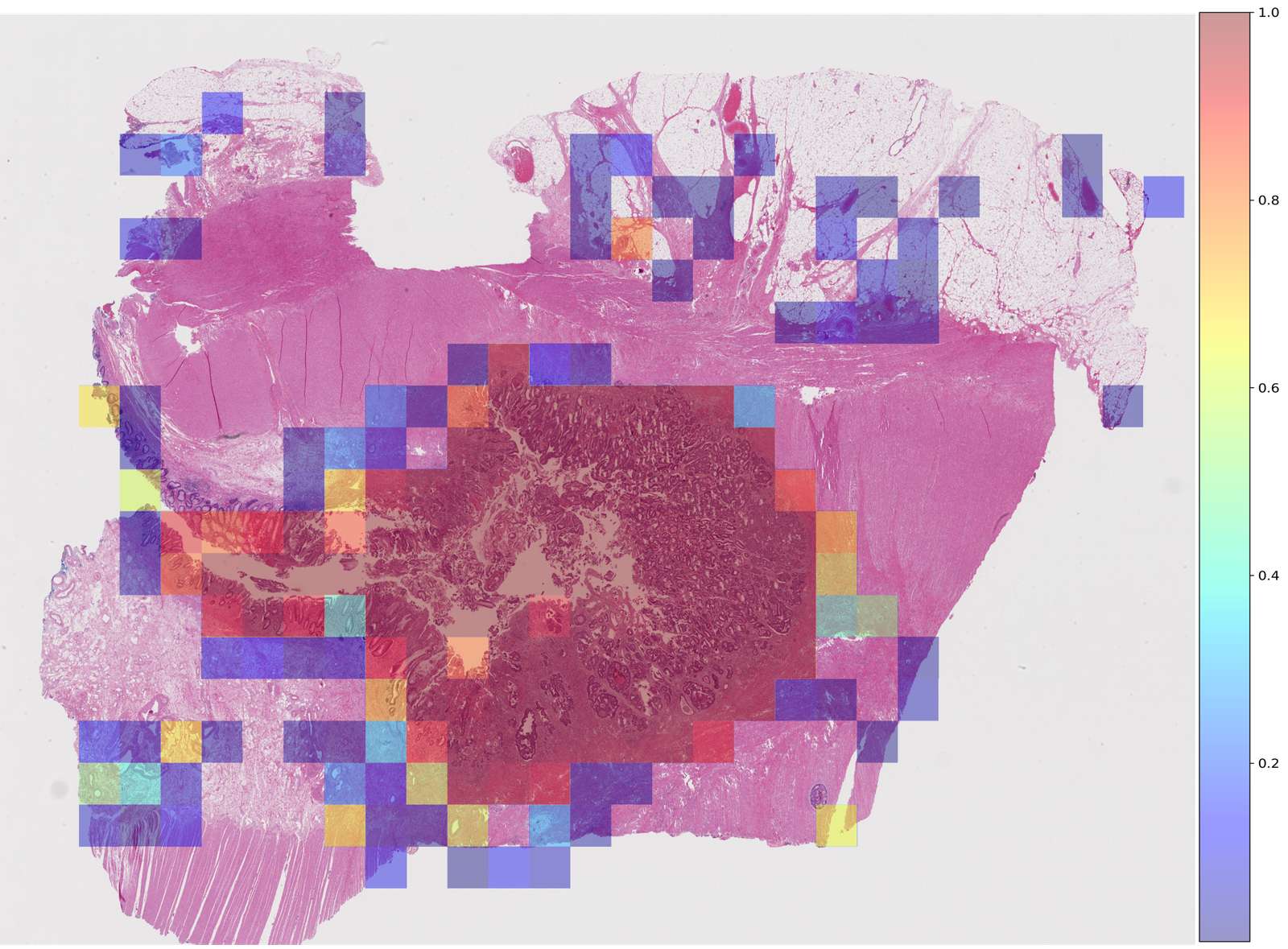}
      \caption{WSI 1: CONCH}
      \label{fig:wsi1-model3}
    \end{subfigure}%
    \label{fig:wsi1}
  \end{subfigure}
  
  \vspace{0.4cm}
  
  % Second row: WSI 2
  \begin{subfigure}{\textwidth}
    \centering
    \begin{subfigure}[b]{0.33\textwidth}
      \includegraphics[width=\textwidth]{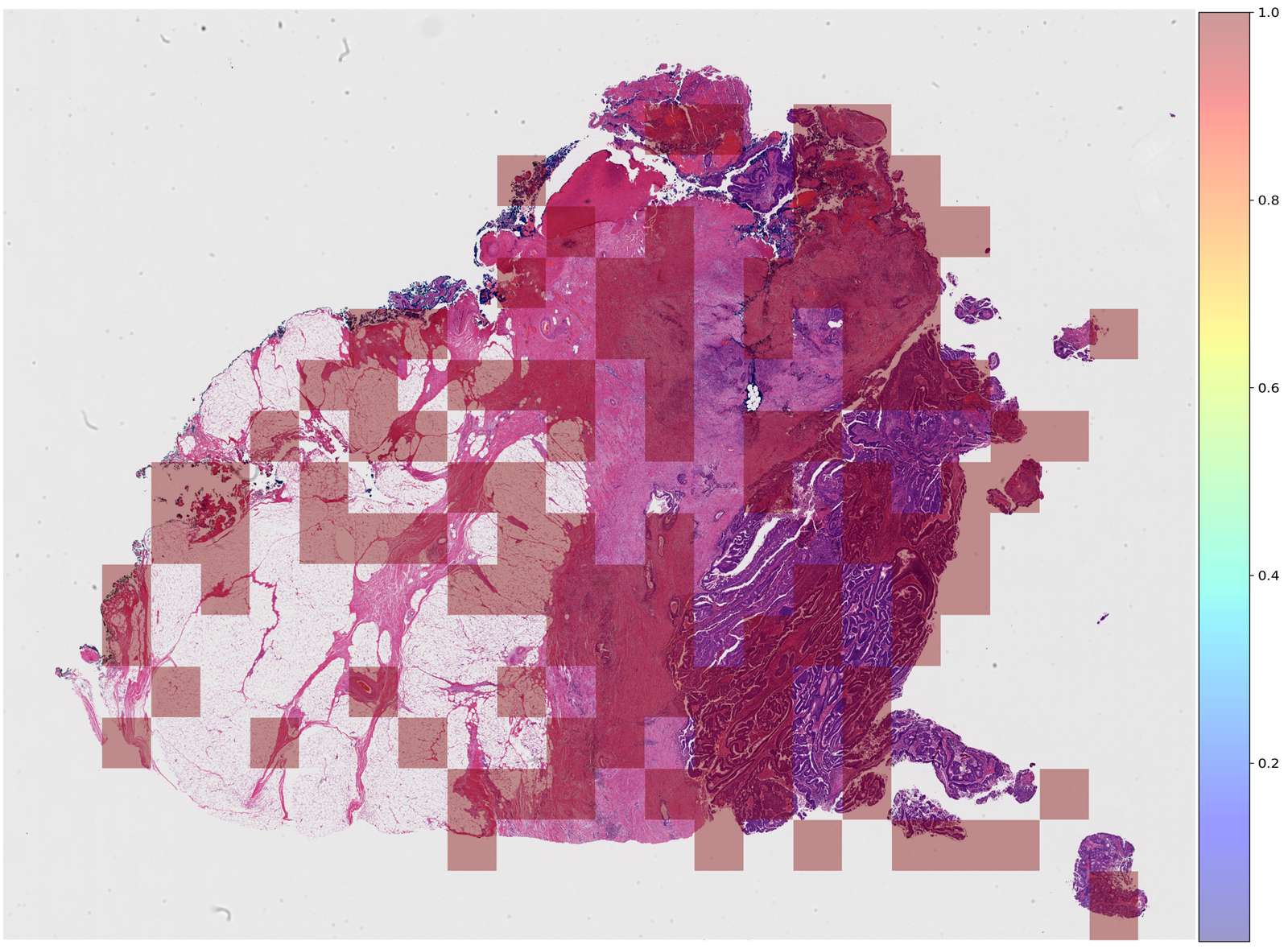}
      \caption{WSI 2: Quilt-Net}
      \label{fig:wsi2-model1}
    \end{subfigure}%
    \hfill
    \begin{subfigure}[b]{0.33\textwidth}
      \includegraphics[width=\textwidth]{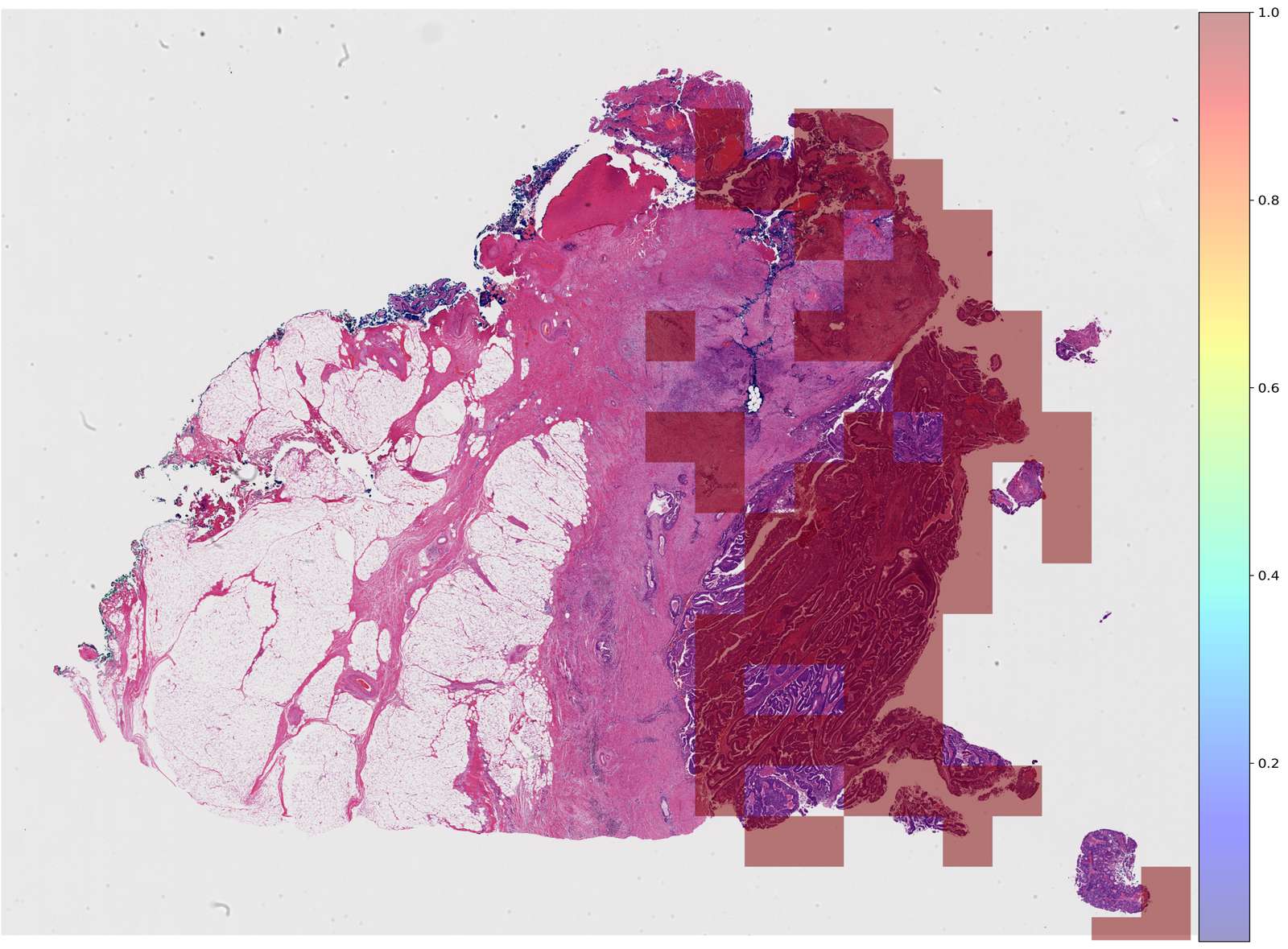}
      \caption{WSI 2: Quilt-LLAVA}
      \label{fig:wsi2-model2}
    \end{subfigure}%
    \hfill
    \begin{subfigure}[b]{0.33\textwidth}
      \includegraphics[width=\textwidth]{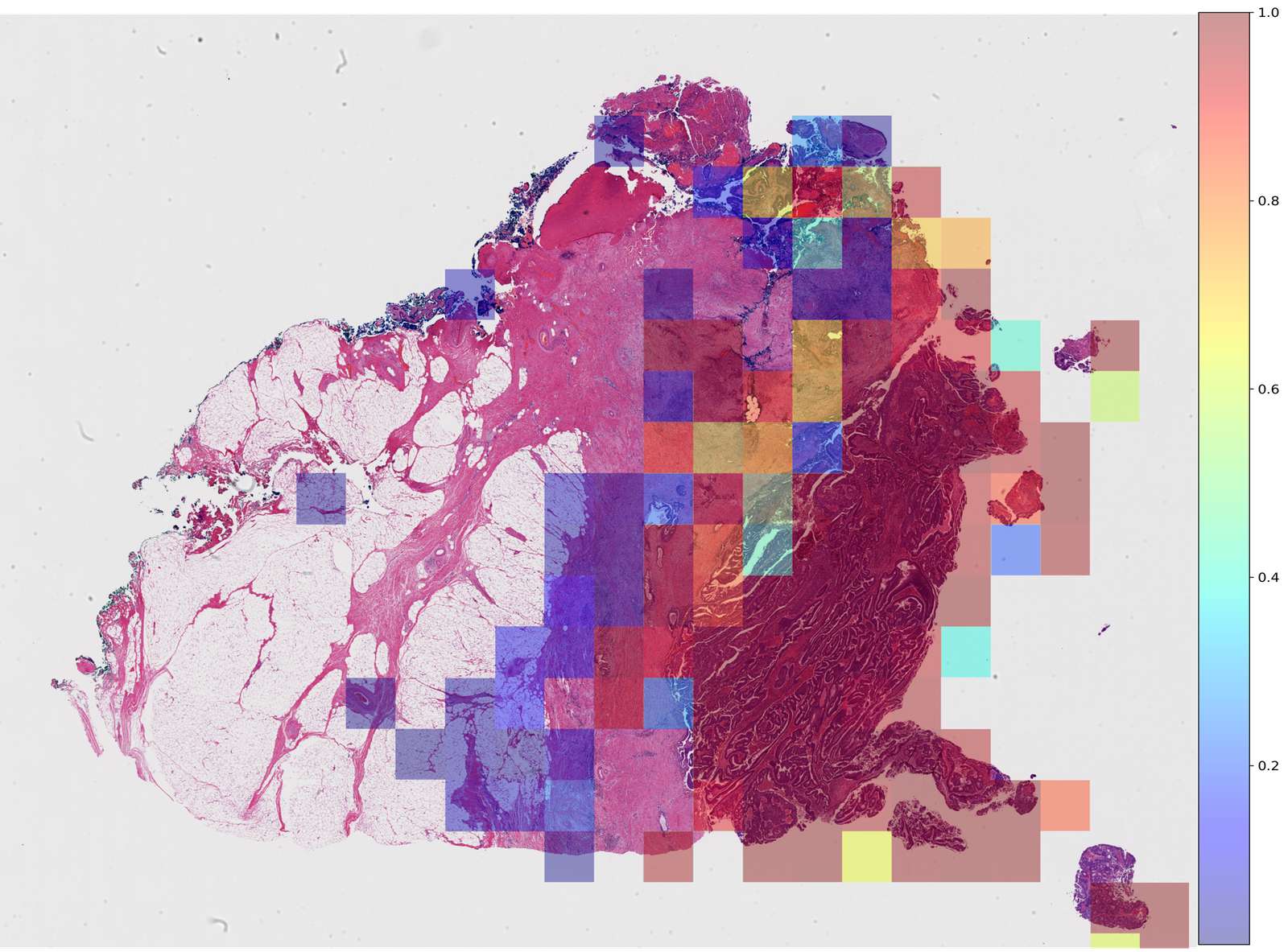}
      \caption{WSI 2: CONCH}
      \label{fig:wsi2-model3}
    \end{subfigure}%
    \label{fig:wsi2}
  \end{subfigure}
  
  \vspace{0.4cm}
  
  % Third row: WSI 4
  \begin{subfigure}{\textwidth}
    \centering
    \begin{subfigure}[b]{0.33\textwidth}
      \includegraphics[width=\textwidth]{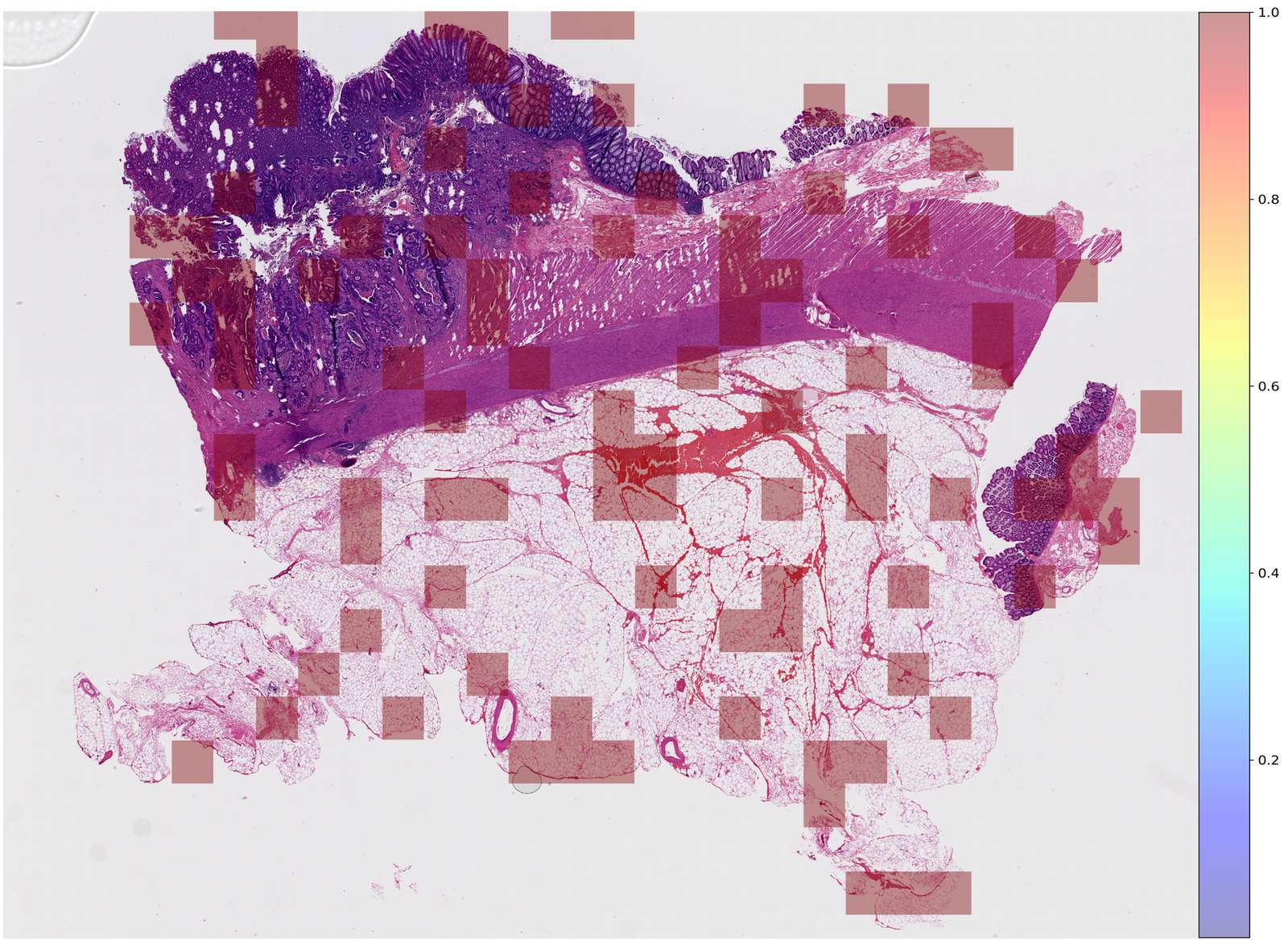}
      \caption{WSI 3: Quilt-Net}
      \label{fig:wsi4-model1}
    \end{subfigure}%
    \hfill
    \begin{subfigure}[b]{0.33\textwidth}
      \includegraphics[width=\textwidth]{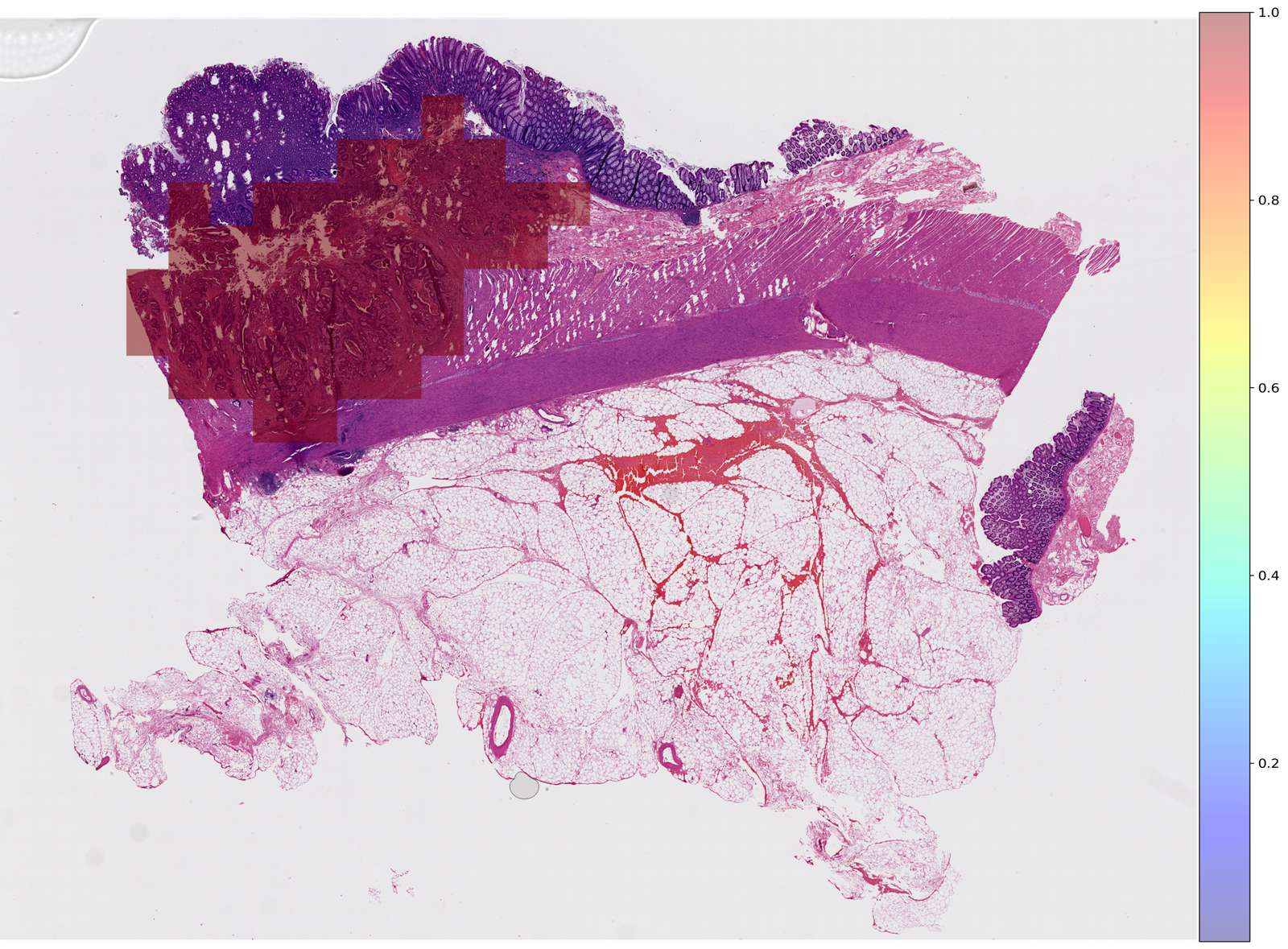}
      \caption{WSI 3: Quilt-LLAVA}
      \label{fig:wsi4-model2}
    \end{subfigure}%
    \hfill
    \begin{subfigure}[b]{0.33\textwidth}
      \includegraphics[width=\textwidth]{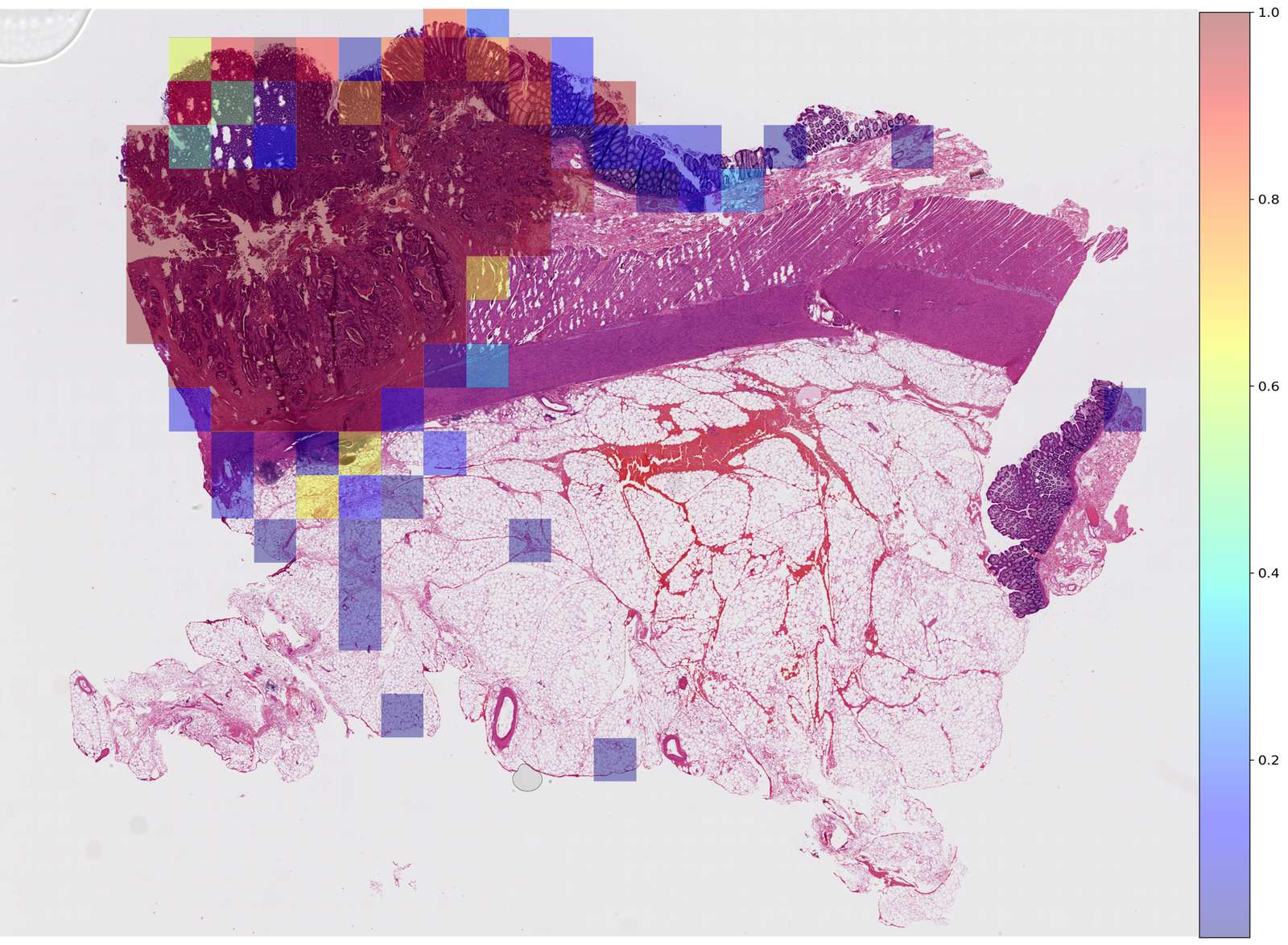}
      \caption{WSI 3: CONCH}
      \label{fig:wsi4-model3}
    \end{subfigure}%
    \label{fig:wsi4}
  \end{subfigure}
  
  \vspace{0.4cm}
  
  \begin{subfigure}{\textwidth}
    \centering
    \begin{subfigure}[b]{0.33\textwidth}
      \includegraphics[width=\textwidth]{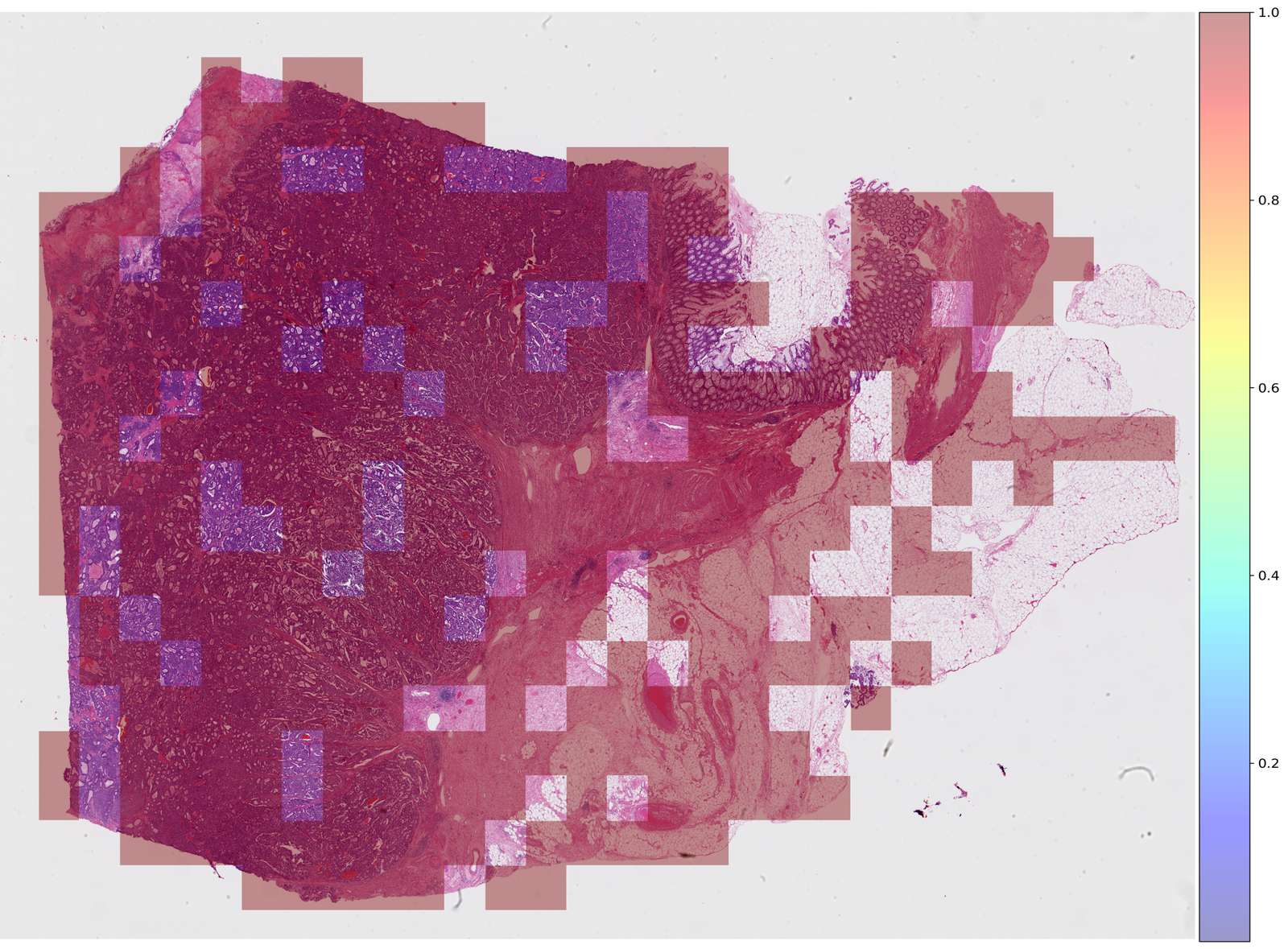}
      \caption{WSI 4: Quilt-Net}
      \label{fig:wsi5-model1}
    \end{subfigure}%
    \hfill
    \begin{subfigure}[b]{0.33\textwidth}
      \includegraphics[width=\textwidth]{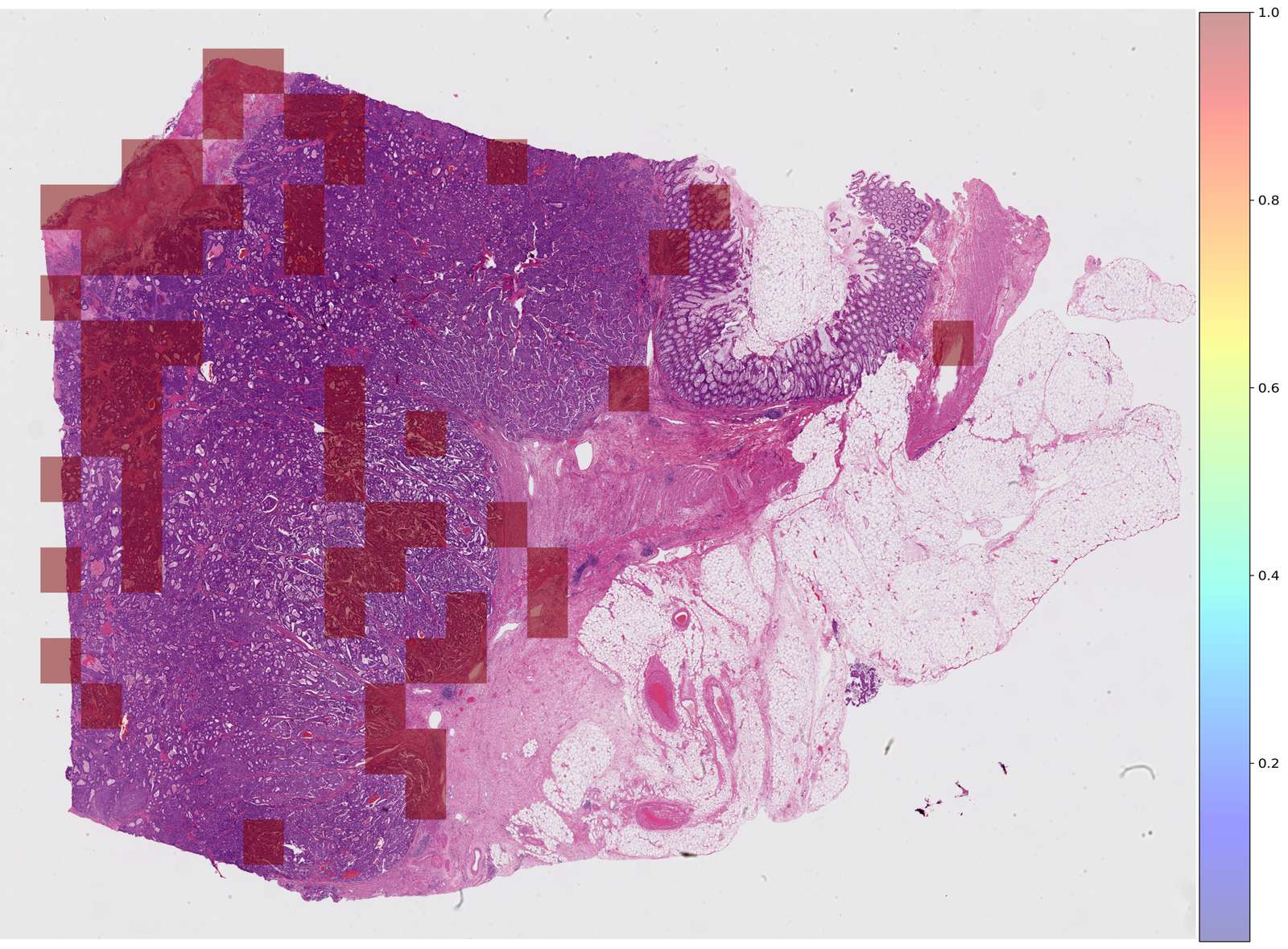}
      \caption{WSI 4: Quilt-LLAVA}
      \label{fig:wsi5-model2}
    \end{subfigure}%
    \hfill
    \begin{subfigure}[b]{0.33\textwidth}
      \includegraphics[width=\textwidth]{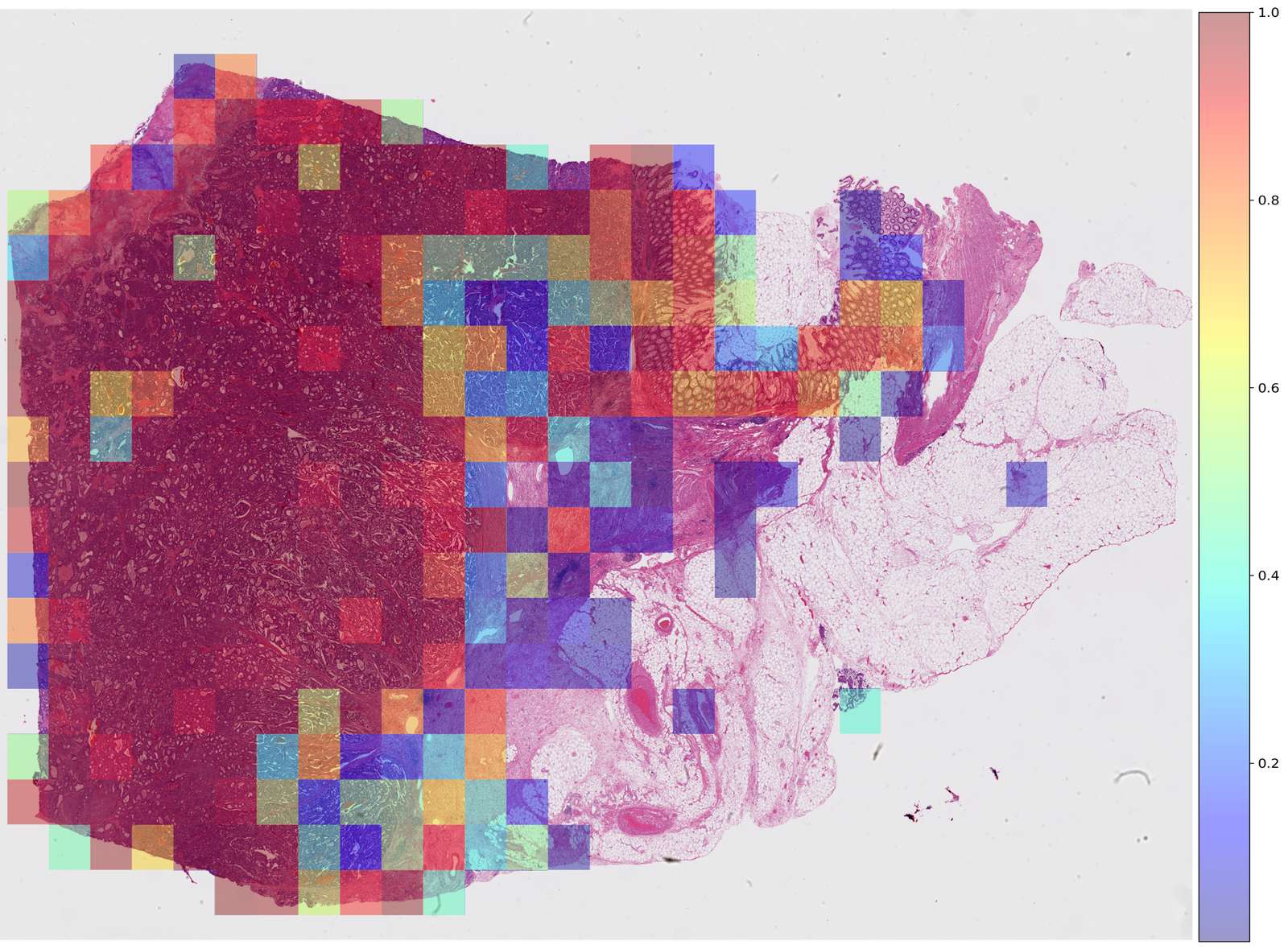}
      \caption{WSI 4: CONCH}
      \label{fig:wsi5-model3}
    \end{subfigure}%
    \label{fig:wsi5}
  \end{subfigure}
  
  \caption{Comparison of different models across multiple WSI samples}
  \label{fig:attention maps}
\end{figure}

Figure \ref{fig:attention maps} presents attention maps generated for histopathological analysis of randomly selected invasive cancer tissue samples. These visualizations represent probability scores assigned at the patch level by the different models to various regions of WSI. The process begins with segmenting each WSI into patches and classifying them as either tissue or background using an in-house CNN. Tissue patches are then processed through the models to obtain probability scores. By default, Quilt-Net and CONCH generate continuous confidence scores, which are used to represent each patch in the WSI and construct a heatmap. In contrast, Quilt-LLAVA does not inherently produce such scores; therefore, binary labels are used to highlight patches or regions identified as invasive. After scoring, the patches are reconstructed with their corresponding probability values to create comprehensive attention maps that highlight regions of interest across the entire WSI, potentially indicating areas of malignancy or specific tissue characteristics. We selected 4 WSIs that cover a diversity of forms of colorectal cancer invading into different levels of depth into the colonic wall.

All three models displayed different attention behavior in the underlying images. Generally, Quilt-Net randomly identified high-attention areas throughout the image, with no significant shift towards areas containing invasive cancer versus areas with cancer. Quilt-LLAVA displayed high-attention patches found within the invasive cancer, but was rather inconsistent in its approach as some areas of the invasive cancer were not highlighted. However, most high attention maps were accurately identified within cancer. CONCH showed the most accurate attention maps of invasive cancer and consistently highlighted its presence throughout the patches. CONCH was more precise in all images but highlighted low-attention areas that were distant from the cancer. CONCH could also highlight at medium-level attention areas of a precursor lesion that is on the verge of becoming cancer and altered tissue areas adjacent to the invasive cancer. Overall, per the review of a board-certified pathologist, CONCH most accurately mimicked the general approach by pathologists in addressing these tissues. Most attention is drawn towards the invasive cancer area, and second-order areas are revised to detect relevant findings, such as precursor lesions and mild changes in the peritumoral area that can be relevant for invasive cancer. Below are detailed analysis on the four WSIs from the board-certified pathologist:

    \textbf{WSI 1} represents a diverticular disease which has progressed into invasive cancer that breaches into the muscularis propria. Quilt-Net targets the whole colon wall with no preference for the invasive cancer versus the non-invaded areas. Quilt-LLAVA targets the invasive cancer and peri-invasive cancer area accurately. CONCH gives high attention at the invasive cancer consistently and highlights at medium attention the precursor area in the epithelium and the affected peri-cancer areas. It notes at low attention the unaffected normal tissue further away.
    \textbf{WSI 2} represents a classical invasive cancer that reached the resection margin and invades into the subserosal connective tissue. Quilt-Net produces randomized high-attention areas throughout the image, Quilt-LLAVA accurately targets the invasive cancer, and CONCH shows high-attention for invasive cancer, medium-attention for the affected pericancer areas, and low-attention to areas without cancer.
    \textbf{WSI 3} represents a classical invasive cancer that is restricted to the muscularis propria, arising from a precursor adenoma. Quilt-Net gives randomized high-attention areas throughout the image, Quilt-LLAVA targets the cancer area while ignoring the precursor lesion, and CONCH targets the cancer area accurately, and at medium- attention the precursor lesion. It further gives low attention to the non-invasive area.
    \textbf{WSI 4} represents a very large cluster of cancer with reactive epithelium at the surface. It invades deeply into the wall into the subserosal connective tissue. Quilt-Net targets the invasive cancer a bit more, but large areas of rather non-invaded tissues. Quilt-LLAVA seems to highlight the cancer, but only in areas that are adjacent to the non-tumoral tissues. CONCH accurately targets the cancer but appears to give low attention to an area of the cancer that is less aggressive while overcalling the reactive epithelium that overlies the cancer.

\section{Conclusion}
This study investigated the impact of prompt engineering and model complexity on the performance of different vision-language models in the domain of computational pathology, namely Quilt-Net, Quilt-LLAVA, and CONCH. Using an in-house digestive dataset of 3,507 whole slide images, we show that prompt design plays a crucial role in model performance, with domain-specific anatomical precision and intermediate level of text details achieving the best performance. Our findings showed that CONCH outperforms the other two models when precise anatomical prompts are used. This proves that effective domain-specific training can matter more than model complexity, since Quilt-LLAVA is a much larger model when compared to CONCH, but is restricted when it comes to model alignment. We also analyze attention maps for sample whole slide images, highlighting cancerous regions detected by the different models. These attention maps are analyzed by certified pathologists, to confirm CONCH's reliability in highlighting diagnostically relevant regions. These findings provide key insights for enhancing VLMs in digital pathology and improving AI-driven diagnostics.

\bibliographystyle{splncs04}
\bibliography{bibliography}

\end{document}